\documentclass{article}

\PassOptionsToPackage{numbers, compress}{natbib}

 \usepackage[preprint]{neurips_2026}

\usepackage[utf8]{inputenc} 
\usepackage[T1]{fontenc}    
\usepackage[hidelinks]{hyperref}       
\usepackage{url}            
\usepackage{booktabs}       
\usepackage{amsfonts}       
\usepackage{nicefrac}       
\usepackage{microtype}      
\usepackage{xcolor}         

\usepackage{amsmath} 
\usepackage{graphicx}
\newcommand{\emoji}[1]{\includegraphics[height=0.8em]{#1}}

\usepackage{tikz}
\usetikzlibrary{arrows.meta, positioning, fit}

\usepackage{amsmath,amsfonts,amssymb}
\DeclareMathAlphabet{\mathbbold}{U}{bbold}{m}{n}

\usepackage{comment}
\usepackage{dirtytalk}

\usepackage{dsfont}

\title{\emoji{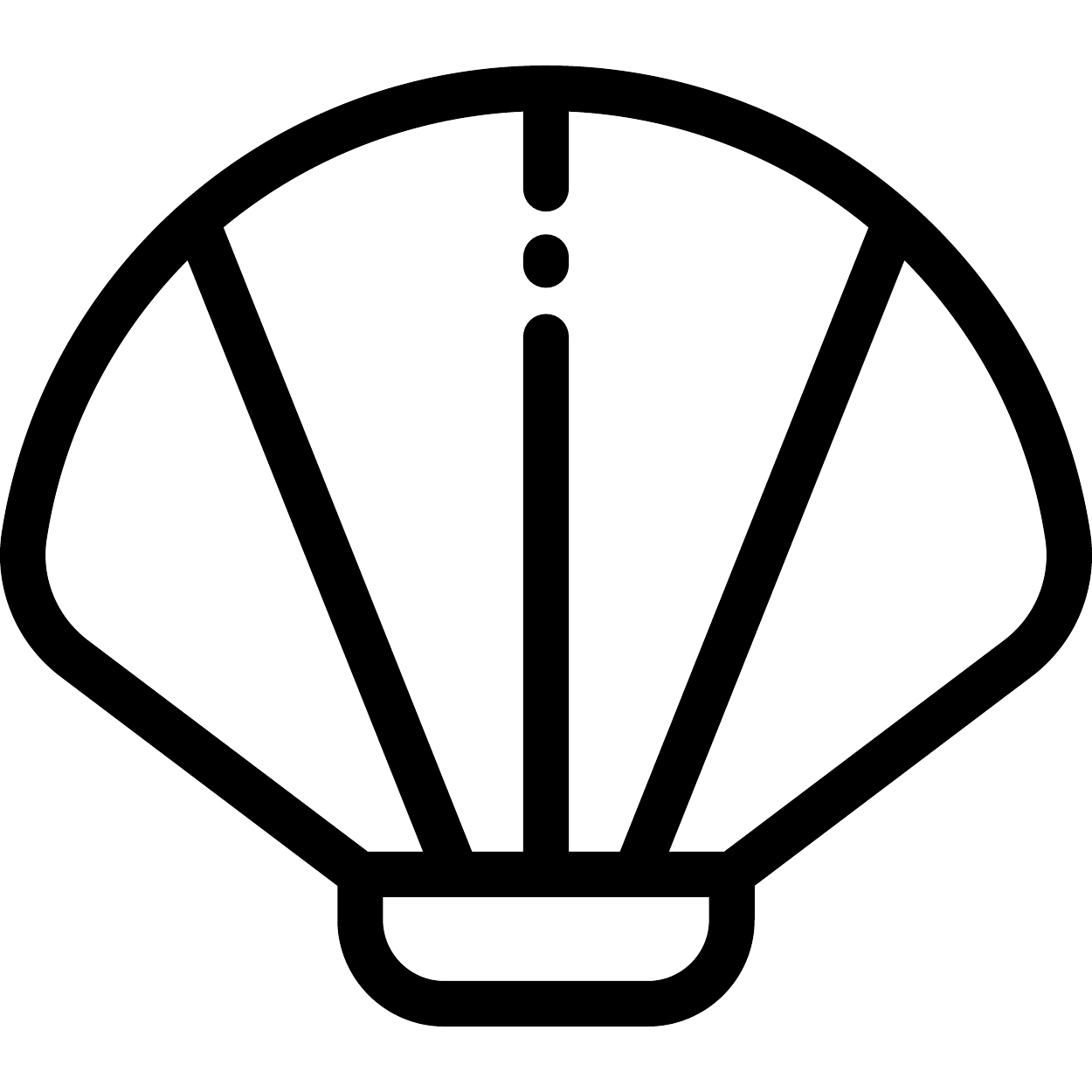} \textsc{Clam}:\\Causal Spatial Disaggregation  \\ to Infer Local Effects From Coarse Data}

\author{%
  Gerrit Gro{\ss}mann \\
  DFKI\thanks{DFKI: Deutsches Forschungszentrum für Künstliche Intelligenz GmbH (German Research Center for Artificial Intelligence). 
  } \\
  Kaiserslautern, Germany \\
  {\scriptsize \texttt{Gerrit.Grossmann@dfki.de} }\\
  \And
  Sumantrak Mukherjee \\
  DFKI\footnotemark[1] \\
  Kaiserslautern, Germany \\
  {\scriptsize \texttt{Sumantrak.Mukherjee@dfki.de} }
  \And
  Sebastian Vollmer \\
  DFKI\footnotemark[1], \;RPTU \\
  Kaiserslautern, Germany \\
  {\scriptsize \texttt{Sebastian.Vollmer@dfki.de} } \\
}

\begin{document}

\maketitle

\begin{abstract}
Learning fine-grained spatial patterns from coarse-resolution data is challenging, especially in causal settings where high-resolution effects must be inferred from aggregated interventions and outcomes. We introduce \textsc{Clam}, a method for estimating localized causal effects from coarse observations by exploiting high-resolution contextual covariates that modulate these effects. By jointly learning the causal mechanism and a disaggregation mapping, \textsc{Clam} captures interactions that are missed when addressing these problems independently. The method supports localized effect estimation, counterfactual reasoning, and principled outcome disaggregation, and reliably captures spatially varying causal effects across diverse settings. This is particularly relevant for applications such as public health and environmental policy, where decisions are made at broad scales despite substantial local heterogeneity. Code is available at \href{https://github.com/gerritgr/clam}{\small\texttt{github.com/gerritgr/clam}}.
\end{abstract}

\section{Introduction}
\begin{figure*}[h!]
    \centering
    \includegraphics[width=0.97\linewidth]{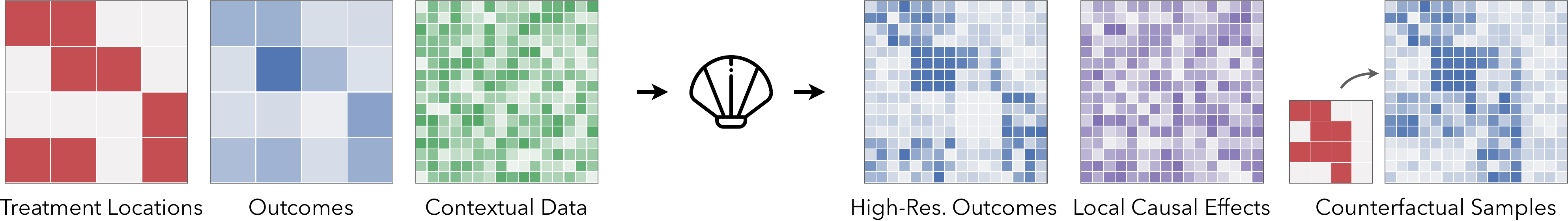}
    \caption{
    \textbf{Motivating example.} 
    \textsc{Clam} takes regional treatments (e.g., a vaccination awareness campaign) and outcomes (e.g., vaccination levels), along with subregional covariates (e.g., demographics), 
    to learn treatment--outcome relationships and estimate local causal effects or sample counterfactuals. }
    \label{fig:motivating_example}
\end{figure*}

Going from high-resolution (HR) to low-resolution (LR) data is typically straightforward, as it can be done via aggregation or pooling functions such as sums or means. These operations generally remove information. In contrast, going from LR to HR data is fundamentally difficult because it requires restoring information that has been lost. This generally necessitates some form of prior knowledge about the HR data. This inverse problem is typically studied under terms such as \emph{spatial disaggregation}~\citep{patil2024systematic}, \emph{downscaling}~\cite{sun2024deep}, \emph{ecological inference}~\cite{schuessler1999ecological}, or \emph{super-resolution}~\citep{anwar2020deep}.  
It is relevant across many fields involving spatio-temporal statistics, including earth science~\cite{kumar2023modern}, public health~\cite{matisziw2008downscaling}, and the social sciences~\cite{ye2016integrating}. 

Causality~\cite{pearl2009causality}, on the other hand, studies the effects of interventions on a system. Its core insight is that statistical associations do not imply causal relationships. In other words, observational patterns may be useful for prediction, but they do not, in general, tell us how outcomes will change under intervention.

Answering causal questions is often attempted using LR data, but this data is frequently insufficient for this~\cite{beckers2019abstracting}. For example, interventions, like vaccination campaigns, might be applied across broad regions, yet their impact could vary significantly at the subregional level.
Surprisingly little effort has been made to properly combine spatial disaggregation with the quest to answer causal questions.

\textbf{Our hypothesis is that causal inference and spatial disaggregation are better
solved together than apart, and that doing so makes questions answerable that neither
can address alone.}

\paragraph{Contribution.}

This work addresses the gap between spatial disaggregation and causal inference by introducing a unified framework grounded in structural causal models (SCMs) and Pearl's do-calculus~\cite{pearl2009causality}. We argue that jointly tackling causal inference and disaggregation is natural in many real-world settings, where interventions are implemented at coarse scales but effects manifest locally. While our primary focus is the formulation of this framework, we illustrate its capabilities and limitations through a series of simulation studies that mimic realistic data-generating processes, as well as a  case study based on real-world data on the impact of heat waves on gun violence.

\textsc{Clam} (\emph{C}ausa\emph{l} dis\emph{a}ggregation \emph{m}ethod) can be summarized as follows: 
We learn a mechanistic function that maps an intervention indicator and contextual information to high-resolution (HR) outcomes. In doing so, we treat HR contextual data (or \emph{covariates}) as an auxiliary variable that implicitly defines a prior over the HR outcome (cf.\ Figure~\ref{fig:motivating_example}). 
The model is trained by aggregating the predicted HR outcomes and enforcing consistency with the observed low-resolution (LR) measurements.
This enables causal effect estimation, counterfactual reasoning, and the computation of HR estimates from LR inputs.

Because no existing method targets this estimand, our contribution is the problem formulation first and the algorithm second, and we prioritize making the assumptions and limits of the setting explicit over benchmarking against methods built for a different task.

\section{Problem Setup\label{problemsetting}}

For notational simplicity and without loss of generality, we assume that all regions share the same subregion structure, interventions are binary, and all variables are scalar rather than vector-valued.

We consider $N$ regions, each divided into $M$ subregions.
The LR \textit{treatment data} is a vector $\widehat{T} \in \{0,1\}^N$, where each entry $\widehat{t}_i$ indicates whether an intervention occurred in region $i$. We assume that, when present, the intervention affects all subregions within that region.

The LR \textit{outcome data} is a vector $\widehat{Y} \in \mathbb{R}^N$, obtained by applying a row-wise aggregation function (e.g., summation or averaging) to an unobserved HR outcome matrix $Y \in \mathbb{R}^{N \times M}$,  where $y_{i,j}$ denotes the outcome value for subregion $j$ within region $i$.
The HR \textit{contextual data} (covariates) is given by a matrix $C \in \mathbb{R}^{N \times M}$, where $c_{i,j}$ denotes the contextual value for subregion $j$ in region $i$.

Finally, we assume that $N$ and $M$ are perfect squares, allowing us to represent the spatial structure as a $\sqrt{N} \times \sqrt{N}$ grid of regions, each containing a $\sqrt{M} \times \sqrt{M}$ grid of subregions (see Figure~\ref{fig:motivating_example} for an example with 
$N=M=16$ and $\sqrt{N} = \sqrt{M} = 4$).

\paragraph{Goal.}
Our goal is to learn a function that estimates local causal effects and to disaggregate coarse outcomes into fine-grained predictions. We formalize this objective in the next section.

\paragraph{Generalizations.}
The setup admits several natural extensions. Variables (such as $c_{ij}$) can be vector-valued, treatments can be continuous instead of binary, and the observed LR treatment $\widehat{T}$ can be interpreted as an aggregation of an underlying HR treatment matrix that varies within a region. The framework also extends to spatio-temporal settings, where $T$, $Y$, and $C$ evolve over time. More generally, additional structure such as hidden confounding, mediation, or non-trivial treatment assignment can be incorporated; we discuss these cases in Section~\ref{sec:motifs}. Naturally, the spatial layout need not follow a regular grid, which we adopt only for illustration.

\paragraph{Causal Assumptions.}
Depending on the experimental setting, different assumptions may hold or be relaxed. We may assume a known causal graph, which constrains the model and reduces the risk of spurious explanations, particularly important when contextual variables are high-dimensional and correlated, but only a subset is causally relevant. We typically assume no hidden confounding, meaning all relevant variables affecting the outcome are observed, allowing interventions to be interpreted causally up to reasonable noise. Causal effect identification also requires variation in interventions across regions. If the aggregation function from subregional to regional outcomes is known (e.g., a mean or sum), disaggregation becomes more tractable.  Moreover, we typically assume that the treatment does not depend on the context values.

\newcommand{\figScale}{0.7}
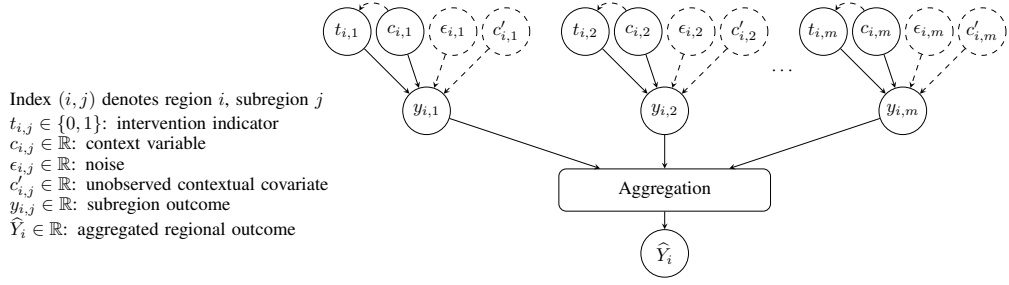
\begin{figure*}[t]
    \centering
    \scalebox{\figScale}{%
    \begin{tikzpicture}[>=stealth,
        uniform circle node/.style={
            circle, draw, minimum size=0.9cm, inner sep=1pt, align=center
        },
        dashed circle node/.style={
            uniform circle node, dashed
        }]
        \node[uniform circle node] (i1) at (0,0) {$t_{i,1}$};
        \node[uniform circle node] (c1) at (1,0) {$c_{i,1}$};
        \node[dashed circle node] (e1) at (2,0) {$\epsilon_{i,1}$};
        \node[dashed circle node] (cp1) at (3,0) {$c'_{i,1}$};
        \node[uniform circle node] (y1) at (1.5,-1.5) {$y_{i,1}$};
        \draw[->] (i1) -- (y1);
        \draw[->] (c1) -- (y1);
        \draw[->,dashed] (e1) -- (y1);
        \draw[->,dashed] (cp1) -- (y1);
        \draw[->,dashed,bend right=60] (c1) to (i1);

        \node[uniform circle node] (i2) at (4.5,0) {$t_{i,2}$};
        \node[uniform circle node] (c2) at (5.5,0) {$c_{i,2}$};
        \node[dashed circle node] (e2) at (6.5,0) {$\epsilon_{i,2}$};
        \node[dashed circle node] (cp2) at (7.5,0) {$c'_{i,2}$};
        \node[uniform circle node] (y2) at (6.0,-1.5) {$y_{i,2}$};
        \draw[->] (i2) -- (y2);
        \draw[->] (c2) -- (y2);
        \draw[->,dashed] (e2) -- (y2);
        \draw[->,dashed] (cp2) -- (y2);
        \draw[->,dashed,bend right=60] (c2) to (i2);

        \node[draw=none] (dots) at (8.25,-0.75) {$\cdots$};

        \node[uniform circle node] (im) at (9,0) {$t_{i,m}$};
        \node[uniform circle node] (cm) at (10,0) {$c_{i,m}$};
        \node[dashed circle node] (em) at (11,0) {$\epsilon_{i,m}$};
        \node[dashed circle node] (cpm) at (12,0) {$c'_{i,m}$};
        \node[uniform circle node] (ym) at (10.5,-1.5) {$y_{i,m}$};
        \draw[->] (im) -- (ym);
        \draw[->] (cm) -- (ym);
        \draw[->,dashed] (em) -- (ym);
        \draw[->,dashed] (cpm) -- (ym);
        \draw[->,dashed,bend right=60] (cm) to (im);

        \node[draw, rectangle, rounded corners,
              minimum width=4cm, minimum height=0.8cm] (agg) at (6.0,-3) {Aggregation};
        \draw[->] (y1) -- (agg);
        \draw[->] (y2) -- (agg);
        \draw[->] (ym) -- (agg);

        \node[uniform circle node] (Yi) at (6.0,-4.2) {$\widehat{Y}_{i}$};
        \draw[->] (agg) -- (Yi);

        \node[draw=none, anchor=north west, align=left, text width=7.2cm]
              (legend) at (-6.5,-1) {%
            Index $(i,j)$ denotes region $i$, subregion $j$\\[2pt]
            $t_{i,j} \in \{0,1\}$: intervention indicator\\
            $c_{i,j} \in \mathbb{R}$: context variable\\
            $\epsilon_{i,j} \in \mathbb{R}$: noise\\
            $c'_{i,j} \in \mathbb{R}$: unobserved contextual covariate\\
            $y_{i,j} \in \mathbb{R}$: subregion outcome\\
            $\widehat{Y}_i \in \mathbb{R}$: aggregated regional outcome};
    \end{tikzpicture}%
    }
    \caption{Causal graph for one region. All functional relations are shared among subregions and regions. Variables could also be vector-valued. Unobserved covariates and a causal link from context to treatment are optional. The aggregation is typically a mean or a sum.}
    \label{fig:scm-design}
\end{figure*}

\section{Our Method: \textsc{Clam}}

Our method combines two insights. First, we represent the causal disaggregation problem using a structural causal model formalism. This formalism encodes assumptions and specifies the missing pieces (functions, parameters). Second, we train an ML model to infer the missing pieces where we use the LR data as a loss signal to estimate the most likely HR data and functional relationships. 

\subsection{Structural Causal Model}
\label{sec:scmmain}
Each region--subregion pair \((i, j)\) is associated with a treatment \(t_{i,j}\), context \(c_{i,j}\), noise \(\epsilon_{i,j}\), and an optional hidden confounder (or non-confounding covariate) \(c'_{i,j}\) (cf.\ Figure \ref{fig:scm-design}). A shared structural function maps these inputs to a subregional outcome \(y_{i,j}\), and then aggregates the subregional outcomes to a regional outcome \(\widehat{Y}_i\):
\begin{equation}
y_{i,j} = f_{\boldsymbol{\theta}}(t_{i,j},\, c_{i,j},\, \epsilon_{i,j},\, c'_{i,j}), \quad 
\widehat{Y}_i = g_{\boldsymbol{\phi}}\bigl(\{y_{i,j}\}_{j=1}^{M}\bigr).
\label{eq:clam-general}
\end{equation}

Each subregion is modeled by a local functional mechanism, with  uncertainty captured via \(\epsilon_{i,j}\). 
Although the functions \(f_{\boldsymbol{\theta}}(\cdot)\) and \(g_{\boldsymbol{\phi}}(\cdot)\) are shared across regions, region-specific dynamics can still be modeled through context variables (e.g., region indicators or positional encodings). The difference between $c'_{i,j}$ and $\epsilon_{i,j}$ is that $\epsilon_{i,j}$ is i.i.d. noise, while $c'_{i,j}$ is structured, for example spatially autocorrelated or shared across observations in time.

In general, \(g_{\boldsymbol{\phi}}(\cdot)\) can be any function parameterized by \(\boldsymbol{\phi}\). For simplicity, we typically assume the aggregation is a sum, the noise is additive, and there are no hidden confounders. Then the model simplifies to:
\begin{equation}
\widehat{Y}_i
=
\sum_{j=1}^{M} \bigl(f_{\boldsymbol{\theta}}(t_{i,j}, c_{i,j}) + \epsilon_{i,j}\bigr) 
=
\sum_{j=1}^{M} f_{\boldsymbol{\theta}}(t_{i,j}, c_{i,j})
+
\sum_{j=1}^{M} \epsilon_{i,j}.
\label{eq:clam-sum}
\end{equation}


If we want to simplify further, we can replace the flexible estimator
$f_{\boldsymbol{\theta}}(\cdot)$ with a linear one, where
$\boldsymbol{\theta}=(\alpha_1,\beta_1,\alpha_2,\beta_2)$. 
Conceptually, we use two linear models, one for treated regions and one for untreated regions, each relating the local context $c_{i,j}$ to the outcome:
\begin{equation}
\widehat{Y}_i
=
\sum_{j=1}^{M}
\bigl(
t_{i,j}\left(\alpha_1 + \beta_1 c_{i,j}\right)
+
(1-t_{i,j})\left(\alpha_2 + \beta_2 c_{i,j}\right)
\bigr)
+
\sum_{j=1}^{M}\epsilon_{i,j}.
\label{eq:clam-linear}
\end{equation}

\subsection{Training}

The model is trained by minimizing the discrepancy between the observed LR outcomes $\widehat{Y}$ and the predicted aggregated outcomes $\mu$, typically using an MSE loss:
$
\mathcal{L}_{\boldsymbol{\theta},\boldsymbol{\phi}} = \frac{1}{N} \sum_{i=1}^N \bigl(\mu_i - \widehat{Y}_i\bigr)^2.
$
Note that we use $\widehat{Y}_i$ for the observed regional outcome and $\mu_i$ for the corresponding prediction, which under zero-mean additive noise is the conditional expectation of the aggregate.

Importantly, the loss is only observed at the aggregated (region) level. The model therefore does not receive direct supervision for individual subregions; instead, it must learn a subregional mechanism that, when combined, is consistent with the observed aggregated outcomes.

When relaxing the assumptions from Section \ref{problemsetting}, training becomes more complicated. For example, if the aggregation function, $g_{\boldsymbol{\phi}} (\cdot)$, is unknown, it must be learned jointly with the local mechanism $f_{\boldsymbol{\theta}}(\cdot)$, which leads to a coupled estimation problem.  In general, $g_{\boldsymbol{\phi}}(\cdot)$ can be implemented using an (order-invariant) set neural network (e.g., \cite{zaheer2017deep}). This is conceptually similar to settings in graph neural networks, where both the message-passing mechanism and the aggregation function are learned \cite{gilmer2017neural}.

Similarly, if treatments are not uniformly applied across subregions, the underlying HR treatment matrix must be inferred jointly during training. More generally, additional latent variables (e.g., unobserved confounders) can be incorporated and estimated within the same framework. This includes uneven delivery, for example a vaccination campaign reaching subregions
to differing degrees.

\subsection{Downstream Applications}

Training yields the predictive mechanism $f_{\boldsymbol{\theta}}(\cdot)$, from which several causal quantities can be derived. The most direct is what we call the \emph{local conditional average treatment effect} (LOCATE), a version of the conditional average treatment effect (CATE) conditioned on subregion-specific factors:
\begin{equation}
\label{eq:simplemaineq}
E_{i,j} = f_{\boldsymbol{\theta}}(1,\, c_{i,j}) - f_{\boldsymbol{\theta}}(0,\, c_{i,j}),
\end{equation}
i.e., the treatment--control difference at the subregional level. Aggregating these quantities yields ATEs for regions, populations, or environments; with continuous treatments, the same mechanism gives conditional dose--response curves. Note that LOCATE is noise-marginalized.

Beyond effect estimation, the framework supports outcome \emph{disaggregation}---mapping coarse regional outcomes back to fine-grained predictions---and counterfactual reasoning, e.g., predicting outcomes under hypothetical intervention placements or alternative contexts. In the simplest case, this amounts to evaluating $f_{\boldsymbol{\theta}}(\cdot)$ under new inputs. For counterfactuals, when the noise term is believed to carry realization-specific information, inferred noise values can be reused to produce sharper, sample-specific counterfactuals rather than population-level averages.

The framework can also recover latent HR treatment locations by representing them as trainable variables (cf.\ Exp.~2 in Appendix~\ref{exp:2details}). With time-series data, \textsc{Clam} can further infer latent variables that modulate causal effects, provided temporal variation supplies enough constraints (cf.\ Exp.~3 in Appendix~\ref{exp:3details}).

\subsection{Identifiability}
\label{sec:identifiabilitySummary}
We do not provide a general identifiability result, and a complete treatment is left as an open problem. Appendix~\ref{app:identifiability} distinguishes causal identification, statistical identification from aggregate observations, and finite-sample recovery by the optimizer, and characterizes several tractable special cases. In short, recovery of the mechanism and recovery of the high-resolution outcome map have to be distinguished, and simple counterexamples show that neither is pinned down without further assumptions. In a noise-free saturated setting with known sum aggregation, the problem reduces to a linear equation system, which makes the required variability in the data explicit through a rank condition. Beyond that, the closest existing result is the consistency analysis of \citet{zhang2020learning}, which applies to an idealized version of our setting. How these results extend to spatially dependent covariates, latent variables, and more general aggregation functions, and what the optimizer actually recovers in finite samples, remain open.

\subsection{Relationship to Causal Deabstraction}

We view \textsc{Clam} as an instance of \emph{causal deabstraction}: recovering a fine-grained causal mechanism from coarse, aggregated observations. While prior work studies when causal structure is preserved under abstraction \cite{beckers2019abstracting,zennaro2022abstraction}, we consider the inverse direction: inferring a high-resolution model that is consistent with observed aggregates. Appendix~\ref{app:CausalDeabstraction} formalizes this view and clarifies the causal semantics. 
Aggregation deletes the pairing between outcomes and subregions. HR covariates and an invariant causal mechanism \cite{peters2016causal,scholkopf2021toward} can partially restore it when covariate compositions vary across regions.

\section{Exploiting Context Across Motifs in the Causal Graph}
\label{sec:motifs}

Throughout the paper, we focus on the simplest causal motif: a treatment $T$ acting on an outcome $Y$, with a contextual covariate $C$ that modulates the effect. Real-world settings can exhibit a much richer variety of causal structures, and the role of high-resolution context, $C$, shifts with the motif: sometimes it must be conditioned on, sometimes ignored, sometimes leveraged for new identification strategies. Do-calculus tells us \emph{whether} an effect is identifiable given the graph, this section asks \emph{how} the LR/HR distinction interacts with that question.
Here, we provide an outlook in which we sketch how HR context can be exploited in common causal graph motifs.

\paragraph{Baseline: $C$ as a Pure Effect Modifier: $T \to Y$, $C \to Y$.}
The default setting throughout this paper is the simplest motif: $T$ and $C$ are independent inputs, both pointing into $Y$, with no edge between them. $C$ is not a confounder, only a covariate that modulates the local effect of $T$, and the backdoor criterion holds trivially and gives rise to the LOCATE in Eq.~\eqref{eq:simplemaineq}. Thus, $C$ provides multiple ``views'' of the same regional outcome and thereby constrains the shared mechanism $f_{\boldsymbol{\theta}}(\cdot)$.

\paragraph{Confounder: $C \to T$, $C \to Y$, $T \to Y$.}
When the context $C$ also drives whether (or by how much) a region is treated, it becomes a confounder. The backdoor criterion then requires conditioning on $C$. Treating $T$ as exogenous yields biased estimates. Modeling the treatment-assignment mechanism at HR is what makes the adjustment correct. Exp.~5 (Appendix~\ref{sec:addexp}) instantiates this motif and shows that ignoring the dependence inflates the MSE on local effects. A concrete real-world example to illustrate this problem is a wildfire-prevention program: forested counties with dry summers are more likely to receive funding, and those same counties also have higher baseline fire incidence, so the program's effect is conflated with the underlying risk.

\paragraph{Collider: $T \to C$, $Y \to C$.}
A collider is a variable influenced by both the treatment and the outcome. Conditioning on such a variable opens a spurious path between $T$ and $Y$, so naïvely passing $C$ as an input to $f_{\boldsymbol{\theta}}(t,c)$ would bias the estimated effect. Nevertheless, HR collider information may still be useful: since $C$ is partly determined by the HR data $y_{i,j}$, it can provide fine-grained information for the reconstruction of $Y$ or serve as a diagnostic check.
For example, consider a regional vaccination campaign ($T$), where the outcome of interest is infection rates ($Y$), and the HR variable $C$ is neighborhood-level pharmacy purchases of rapid tests. The campaign may increase public attention and test purchases through its messaging, while actual local infections also increase demand for rapid tests. 

A possible way to exploit this structure is to invert the usual roles: instead of using $C$ as a covariate of the outcome, we treat $C$ itself as the HR target and learn a local mechanism $f^{\mathrm{inv}}(t,y) = c$ that maps treatment and outcome to the collider. Under this setup, \textsc{Clam}'s aggregation-consistency principle still applies, but it is used to jointly learn the mechanism and the disaggregation of $T$ and $Y$ against the observed HR values of $C$.

\paragraph{Mediator: $T \to C$, $C \to Y$, $T \to Y$.} A mediator \(C\) lies on the causal pathway from \(T\) to \(Y\), so part of the treatment effect is transmitted through \(C\) rather than acting directly. Na\"ive estimation that conditions on \(C\) cannot separate the direct effect of \(T\) from the indirect effect through \(C\). When the high-resolution mediator is observed, one can instead jointly learn the treatment-to-mediator, mediator-to-outcome, and direct treatment-to-outcome mechanisms. Only the two outcome-side mechanisms rely on the aggregated regional observations, since the treatment-to-mediator mechanism is supervised directly by the observed \(c_{i,j}\). This enables a decomposition of the learned effect into a direct component \(T \to Y\) and an indirect component \(T \to C \to Y\). A concrete example is a regional advertising campaign observed at low resolution, whose effect on regional sales, also observed at low resolution, is partly direct and partly mediated by word-of-mouth dynamics reflected in local online discourse observed at high resolution.

\paragraph{Instrumental variable: $C \to T$, $T \to Y$, $U \to T$, $U \to Y$.} If a hidden confounder links treatment and outcome, but the HR covariate acts on the outcome only through the treatment, that covariate may play the role of an instrument. An HR instrument is unusual but plausible. Consider subregional weather ($C$, HR), which influences whether a vaccination campaign is deployed ($T$, observed at LR), but does not affect the infection rate ($Y$, observed at LR) in any other way. One possible extension of \textsc{Clam} would then be to disaggregate treatment and outcome jointly, as $y_{i,j} = f_2(f_1(c_{i,j}))$, where $f_1(\cdot)$ maps weather to local deployment and is trained against the LR treatment totals, and $f_2(\cdot)$ maps deployment to local infections and is trained against the LR outcome totals. In such a construction, $f_2(\cdot)$ would only see the deployment predicted from the weather, which may help isolate the component of treatment variation attributable to the instrument rather than the hidden confounder.

\paragraph{Spill-over.}
A subregional outcome may depend on neighboring treatments and contexts: interventions can spill over across space. The local mechanism is then no longer purely local, and aggregated observations alone cannot distinguish a strong local effect from a weaker effect that diffuses outward. HR context can help by exposing spatial structure. Technically, one can replace the pointwise $f_{\boldsymbol{\theta}}(\cdot)$ with a grid-based model. In principle, the model can learn both direct local effects and spill-over patterns from aggregated outcomes. A concrete example is a driving ban in one district that displaces traffic into adjacent ones, so regional air quality depends on both the ban location and surrounding road and vegetation patterns.

\paragraph{Time-Varying Confounding.}
In spatiotemporal settings, confounders may evolve over time and be shaped by past treatments. Present context can then no longer be treated as a fixed pre-treatment covariate, because it may already encode earlier interventions. This makes estimation harder: naïve adjustment can block part of the causal pathway or introduce bias \cite{liley2021model}. \textsc{Clam} can be extended by indexing $f_{\boldsymbol{\theta}}(\cdot)$ over time and by explicitly modeling how past treatments, outcomes, and HR covariates affect the current state. Exp.~3 already uses temporal variation to recover a stable latent spatial factor; a natural next experiment would make this factor dynamic and treatment-dependent. For example, neighborhoods with high prior infection rates may receive additional public-health resources, which then change later vulnerability and influence future allocation decisions.

\paragraph{Summary.}
Across all motifs, the role of HR data can be summarized as follows: it provides \emph{variation within regions} that translates into observable differences at the aggregated level. This variation can be exploited to constrain and identify fine-grained causal mechanisms, but only when the causal graph permits it.

More concretely:
(i) for effect modifiers and confounders, HR structure is highly informative and can be directly leveraged for identification and adjustment;
(ii) for mediators, it remains informative but requires explicitly modeling the underlying causal pathways;
(iii) for colliders, it can be actively misleading if the framework is not changed accordingly.

\textsc{Clam} does not replace causal identification via do-calculus: the causal graph determines whether a target effect is identifiable in principle, while \textsc{Clam} estimates the corresponding fine-grained structure from aggregated data. Identifiability in principle, however, does not guarantee recovery in practice, which further requires sufficient variability in the HR covariates, a well-behaved optimization problem, and effective training of the chosen function class. See Appendix~\ref{app:identifiability} and Section~\ref{sec:limitations}.

\section{Related Work}
\label{sec:relatedwork}

\textsc{Clam} sits at the intersection of multiple research strands; an extended discussion is provided in Appendix~\ref{app:relatedwork}.

\paragraph{Spatial Disaggregation and Downscaling.}  
A long line of work in statistical downscaling and spatial disaggregation infers high-resolution estimates from coarse data using interpolation, regression, or learning-based techniques \cite{patil2024systematic,sun2024deep,maraun2019statistical,chau2021deconditional}. These methods are predominantly associational, recent extensions integrate causally relevant predictors \cite{vazquez2022effectiveness,dutta2020identification} but still rely on relatively fine-grained information and do not target interventional or counterfactual queries.

\paragraph{Causal Inference under Aggregation and Abstraction.}
Reasoning causally with aggregated data is well known to be hazardous: results on the ecological fallacy show that aggregate statistics may obscure or invert relationships at the unit level \cite{freedman1999ecological,rogers1991aggregation,weinstein2026hierarchical,li2026spatio}. Work on causal abstraction formalizes when mappings between models at different granularities preserve interventional or counterfactual semantics \cite{beckers2019abstracting,beckers2020approximate,zennaro2022abstraction}, and spatiotemporal causal inference \cite{christiansen2022toward,zhou2024estimating,mukaigawara2025spatiotemporal,song2024bipartite,oprescu2026gst,ninad2025causal,balkus2024causal} addresses related challenges with different data regimes. None of these directly target the joint problem of disaggregating outcomes \emph{and} estimating local causal effects from coarse interventions.

\paragraph{Invariant Mechanisms, GNNs, and Hierarchical Models.}
Conceptually, \textsc{Clam} relies on the principle of \emph{invariant causal mechanisms} \cite{peters2016causal,scholkopf2021toward,heinze2018invariant,rojas2018invariant}: the local mechanism $f_{\boldsymbol{\theta}}(\cdot)$ is shared across subregions while contextual covariates drive heterogeneity. This turns aggregation into a constraint and connects our architecture to message-passing GNNs \cite{gilmer2017neural} and Deep Sets \cite{zaheer2017deep}, as well as to hierarchical Bayesian models that pool information across groups \cite{feller2015hierarchical,dias2013hierarchical}. It also places \textsc{Clam} in the setting of learning from aggregate observations \cite{zhang2020learning}, whose training loss coincides with ours under mean aggregation (cf.\ Appendix~\ref{app:identifiability}).

\section{Experiments}
\label{sec:experiments}

\begin{figure*}[t]
  \centering
  \includegraphics[width=0.97\textwidth]{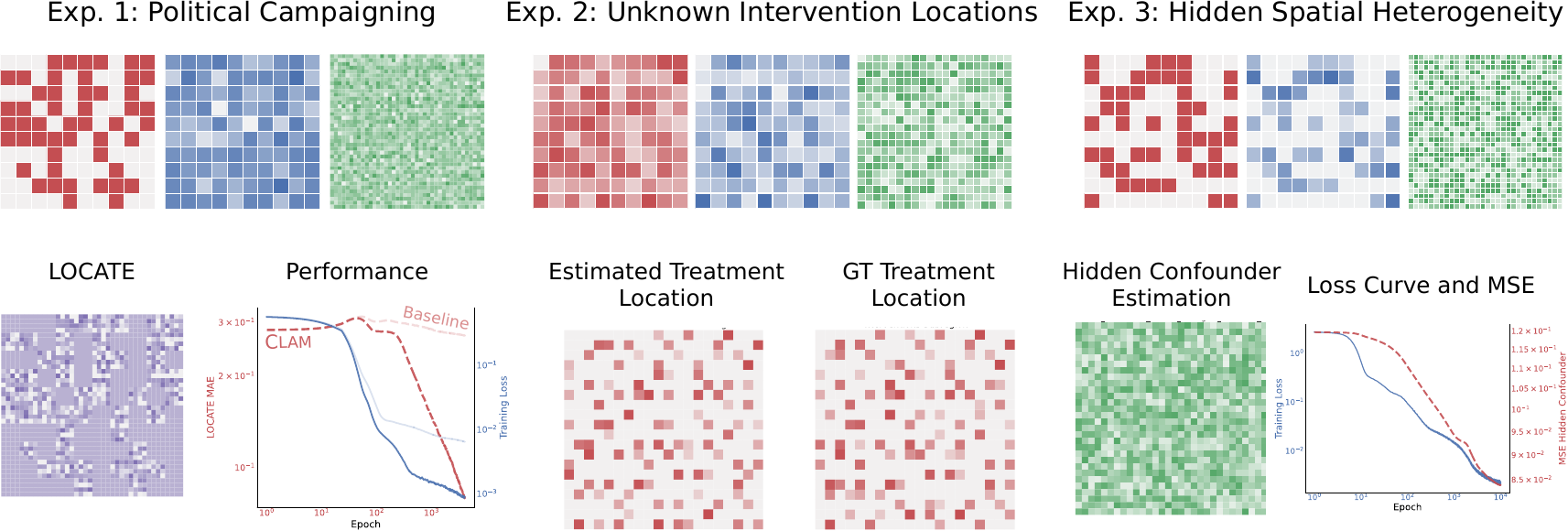}
  \caption{
    \textbf{Results Overview.}
    \textbf{Top:} Input of LR intervention locations (\textcolor[rgb]{0.768,0.305,0.321}{red}),
    LR outcomes (\textcolor[rgb]{0.298,0.447,0.690}{blue}),
    and HR context (\textcolor[rgb]{0.333,0.658,0.407}{green}).
    \textbf{Bottom:} Selected inference results.
  }
  \label{fig:results}
\end{figure*}

We evaluate \textsc{Clam} on three synthetic studies and one real-world study. In the synthetic studies the ground truth mechanism is available isolate distinct capabilities of the framework under controlled conditions: heterogeneous local-effect recovery, latent intervention localization, and the reconstruction of unobserved covariates. The real-world study then probes the method on U.S.\ gun-violence data, where we assume that only coarse regional outcomes are observed.

Full experimental details, per-experiment figures, and additional analyses are provided in Appendices~\ref{app:expdetails} and~\ref{app:realworld}, with the ablation study included in Appendix~\ref{app:expdetails}.
Two additional studies---covering confounded treatment allocation and an unknown aggregation function---are deferred to Appendix~\ref{sec:addexp}.
Our focus is qualitative: we aim to characterize the behavior of the method.
Figure~\ref{fig:results} provides an overview of the inputs and headline results. Code that runs off-the-shelf on Google Colab is provided.

\subsection{Synthetic Studies}

\paragraph{Exp.~1: Political Campaigning.}
We first ask whether \textsc{Clam} can recover heterogeneous local causal effects from aggregated outcomes when the effect of treatment depends on subregional context. We model a binary regional intervention (campaign spending) whose impact varies with subregional wealth (z-scored income values). Regional-level covariates largely remain uninformative by construction. The reported outcome is the candidate's relative improvement, observed only at the regional level. We compare against a naïve \textbf{baseline} (Figure~\ref{fig:results}, with confidence intervals in the Appendix) that applies a uniform disaggregation mechanism to the regional outcome and then learns the causal mechanism.

In the limiting case where all regional outcomes are identical, uniform, bicubic, and ecological-regression \citep{goodman1953ecological} baselines necessarily reach the same LOCATE error floor: without informative HR context, they cannot distinguish subregions and can at best assign the regional average effect. Exp.~1 is constructed close to this regime, with only minimal regional outcome variation, so explicitly evaluating these additional baselines provides little additional information. \textsc{Clam} operates below this floor because the shared mechanism ties each outcome to the context of its subregion. This is the gain hypothesized in the introduction, where treating disaggregation and causal inference separately caps what is achievable.

We use a simple neural network to estimate the causal mechanism. \textsc{Clam} accurately recovers the underlying subregional causal effects, with the MAE to the ground-truth LOCATE matrix (cf.\ Eq.~\eqref{eq:simplemaineq}) decreasing alongside the training loss. The learned model supports coherent counterfactual reasoning at both regional and subregional resolutions and correctly identifies regions whose outcome differences arise purely from contextual heterogeneity rather than mean effects. Full setup and per-subregion visualizations are given in Appendix~\ref{exp:1details} (Figure~\ref{fig:results_exp1}).

\paragraph{Exp.~2: Public School Funding.}
Next, we test whether \textsc{Clam} can recover \emph{latent intervention locations} jointly with local causal effects when only aggregated intervention totals and outcomes are observed. Each of the $30 \times 30$ regions consists of $4\times 4$ subregions and receives binary funding in either no subregion or exactly one unobserved subregion. Treatment effects vary with subregional socioeconomic status, and the outcome is education-score improvement. We implement $f_\theta(\cdot)$ using either a parametric function or a small neural network. In both cases, \textsc{Clam} jointly identifies the treated subregions and accurately recovers the underlying causal-effect function. To sample a treated subregion, we use the common Gumbel-Max trick. Details are provided in Appendix~\ref{exp:2details} and Figure~\ref{fig:exp2details}.

\paragraph{Exp.~3: Spatiotemporal Effects of Heat Waves.}
Finally, we ask whether \textsc{Clam} can reconstruct an \emph{unobserved spatial effect modifier} from aggregated outcomes in a spatiotemporal setting. Heat waves negatively affect school performance, with heterogeneous impact depending on parental education (observed) and vegetation coverage (unobserved), the latter mitigating heat exposure. Only region-level outcomes are observed over time. Under a correctly specified linear mechanism, both the regional prediction loss and the MSE converge rapidly, and the estimated disaggregation of the vegetation is visually indistinguishable from the ground-truth vegetation map. With a flexible MLP, \textsc{Clam} still recovers meaningful spatial structure but with substantially higher error. This reflects an identifiability trade-off whereby expressive models can absorb transformations of the vegetation while preserving aggregated predictions. Details are provided in Appendix~\ref{exp:3details} (Figure~\ref{fig:results_exp3}).

\paragraph{Takeaways.}
Across the three studies, \textsc{Clam} performs well in the synthetic setting, particularly for estimating causal effects. The recovery of latent variables, such as intervention locations or unobserved effect modifiers, depends on the specific experiment. Overfitting may occur, when large amounts of data is available.

\subsection{Semi-Synthetic Case Study With Real-World Data: Heat and Gun Violence}
\label{sec:real_world_experiment}

\begin{figure}[t]
    \centering
    \includegraphics[width=0.7\linewidth]{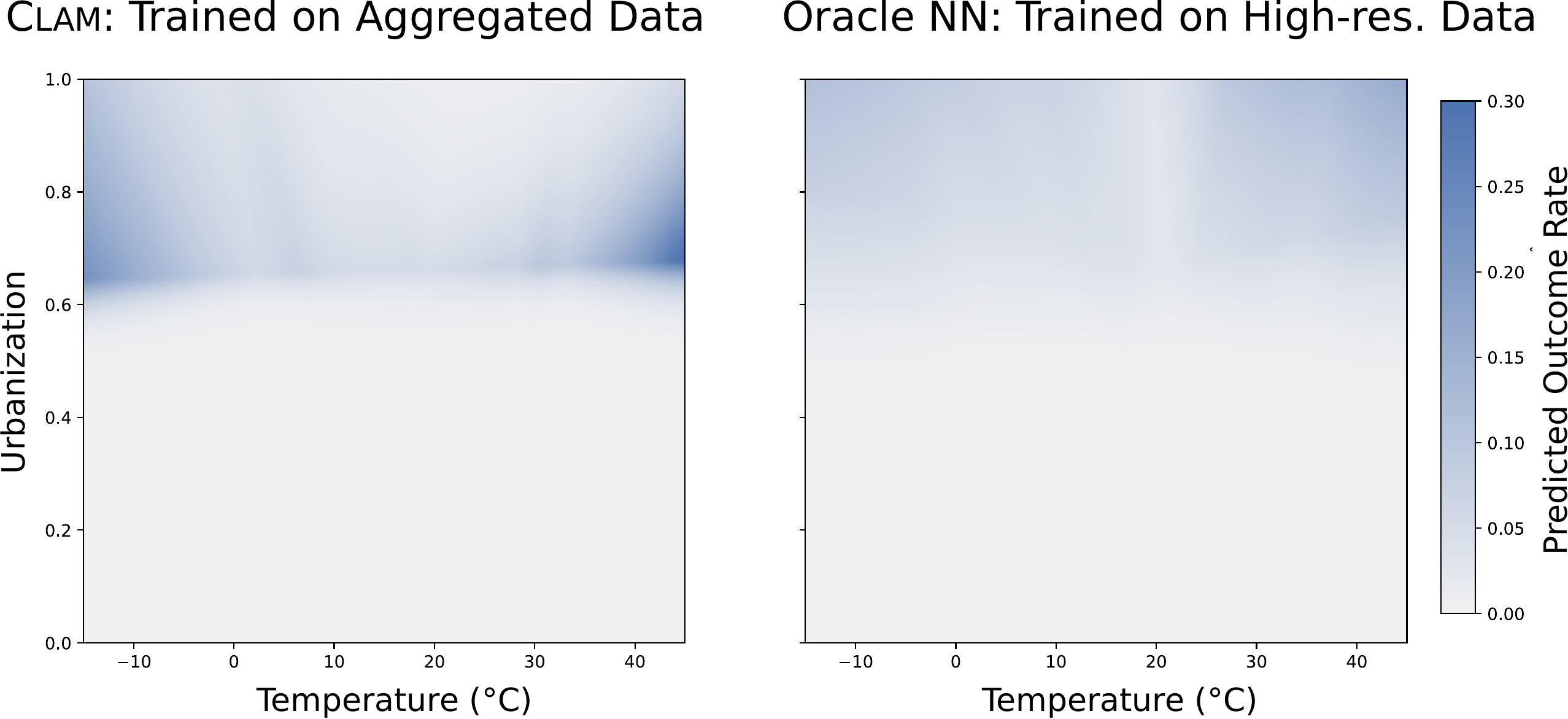}
    \caption{Side-by-side comparison of the learned rate functions $f_{\boldsymbol{\theta}}(\text{temp}, \text{urban})$ for the disaggregation from \textsc{Clam} (left) and the oracle (a NN with the ground truth subregional data; right). Predictions are generated across a temperature range of $[-15,45]^\circ\mathrm{C}$ and an urbanization scale of $[0,1]$. 
    }
    \label{fig:comparison_heatmap}
\end{figure}

We evaluate \textsc{Clam} on U.S.\ gun-violence data from the Gun Violence Archive (GVA) for 2022--2023~\cite{gunviolencearchive2015}, motivated by prior evidence linking higher ambient temperatures to increased firearm violence~\cite{lyons2022analysis}. We treat temperature as a continuous exposure. Urbanization is the high-resolution (HR) contextual covariate, and gun-incident counts are observed at weekly resolution.
A core difficulty in evaluating disaggregation methods on real data is the absence of ground truth at the fine scale. We work around this by starting from data that is natively HR (incident counts at the cell level) and \emph{artificially} aggregating it into LR regional counts, which serve as the inputs to \textsc{Clam}. To construct a high-quality reference \textbf{baseline}, the original HR counts are only used for fitting an \say{oracle} that we use for evaluation.

We use $N=100$ regions and $M=100$ cells per region, indexed by region $i$, subregion $j$, and week $w$. Let $Y^{\mathrm{LR}}_{i,w}$ be the aggregated regional count in region $i$ during week $w \leq 52$, $c_{i,j}$ the HR urbanization covariate, and $\widetilde{t}_{i,w}$ the standardized regional temperature. We model subregion-level rates as $\lambda_{i,j,w}=f_{\boldsymbol{\theta}}(\widetilde{t}_{i,w},c_{i,j})$ and aggregate them as $\mu_{i,w}=\sum_{j=1}^{M}\lambda_{i,j,w}$, with $Y^{\mathrm{LR}}_{i,w}\sim\mathrm{Poisson}(\mu_{i,w})$. Training minimizes the regional Poisson negative log-likelihood.
Figure~\ref{fig:comparison_heatmap} shows that \textsc{Clam} recovers a qualitatively similar temperature--urbanization interaction to the oracle. The spatial comparison in Appendix Figure~\ref{fig:spatial_comparison_heatmap} shows the same pattern: plausible structure, but inflated and more diffuse local intensities, reflecting an aggregation-induced identifiability gap.

\section{Limitations}
\label{sec:limitations}

\textsc{Clam} relies on assumptions that are necessary given the nature of inferring
fine-grained causal effects from aggregated data, but are nonetheless restrictive.
We discuss the main limitations here.

\paragraph{Prerequisite: Do-Calculus Identifiability.}
\textsc{Clam} is not a substitute for causal identification. The method should only be applied in settings where do-calculus (applied to a known or expert-specified causal graph) establishes that the causal effect is identifiable in principle. If the causal graph is unknown or the required identification conditions fail (e.g., the backdoor criterion is violated by unmodeled confounders), \textsc{Clam} will fit a model to the data that does not carry a causal interpretation.

\paragraph{No Identification Guarantee from Optimization Alone.}
Even when do-calculus certifies identifiability in principle, we do not provide a formal guarantee
that our optimization procedure recovers the true causal effect. The training objective is
non-convex, and the best we can reasonably claim is that, under good initialization and
sufficient covariate diversity, gradient descent finds a local minimum whose value is close to
the true causal effect. It is easy to come up with pathological cases in which different fine-grained disaggregations
are equally consistent with the observed regional outcomes. How to detect or avoid them in practice, remains an open question.

\paragraph{Assumptions and Their Scope.} Successful recovery of the local mechanism generally relies on (i) an invariant causal mechanism across subregions, (ii) sufficient diversity in subregional covariate distributions across regions, (iii) no hidden confounding, or its explicit modeling, and (iv) a known or sufficiently constrained aggregation function.

The invariance assumption (i) is less restrictive than it may appear: although $f_{\boldsymbol{\theta}}(\cdot)$ is shared across regions and subregions, region-specific dynamics can still be captured through contextual covariates such as region indicators or positional encodings. Apparent heterogeneity across regions is absorbed by the context, yielding a \emph{conditionally} invariant model. The stronger requirement is that the contextual covariates provide sufficient variation to constrain the shared mechanism and capture sources of heterogeneity that are correlated with the treatment. Unobserved factors that are uncorrelated with the treatment can effectively act as additional noise: they make recovery harder but need not bias the estimated effect. Unobserved confounders, by contrast, will generally lead to incorrect causal estimates unless they are explicitly modeled. The aggregation function (iv) need not be perfectly known in advance: as demonstrated in Exp.~4 (Appendix~\ref{sec:addexp}), it can be jointly learned alongside the local mechanism when it is sufficiently parameterized and the available variation allows the two components to be distinguished.

\paragraph{Identifiability Gap and Latent Variable Recovery.} Even when all assumptions hold, aggregation induces an intrinsic identifiability gap: matching aggregated predictions to aggregated observations may admit multiple equally valid solutions. Our real-world study illustrates this, with \textsc{Clam} systematically inflating local effect magnitudes relative to the cell-supervised oracle. The problem is amplified when latent variables are jointly inferred (e.g., intervention locations in Exp.~2, unobserved covariates in Exp.~3), since different inferred values can each correspond to a mechanism that explains the data equally well. We discuss this further in Appendix~\ref{app:identifiability} and regard a rigorous treatment, together with practical diagnostics, as the most important direction for future work.

\paragraph{Limited Real-World Validation.} Our evaluation is primarily qualitative and relies substantially on synthetic studies, where the data-generating assumptions are known and can be matched to the model. The real-world study uses a subregion-supervised oracle as a reference rather than a true causal benchmark. However, the identified causal effect of temperature on gun-violence intensity is also supported by prior literature \cite{lyons2022analysis}. Robustness to noise, measurement error, and treatment--context dependence is explored in the additional experiments, but a more systematic benchmarking remains an important direction for future work.

\section{Conclusions and Future Work}
The primary contribution of this work is not a single algorithm but the articulation of a \emph{problem class}: estimating high-resolution causal effects from low-resolution interventions and outcomes by exploiting high-resolution context. We frame this as \emph{causal deabstraction}, inverting the standard notion of causal abstraction and recasting spatial aggregation as a constraint satisfaction problem, thereby bridging structural causal models and spatial statistics. In many real-world settings such as public health, environmental policy, and education, spatial disaggregation and causal estimation are inseparable, since interventions are implemented at coarse scales but their effects manifest locally. Treating them jointly within a structural causal model makes assumptions explicit, clarifies what is and is not identifiable, and turns causal inference into a tractable optimization problem.

\textsc{Clam} is one concrete instantiation of this framework. It explicitly adjusts for subregional treatment-context confounding using only coarse regional supervision, a challenging problem in ecological inference, and jointly learns latent high-resolution variables such as exact intervention locations and latent spatial effect modifiers alongside the causal mechanism. Across synthetic and real-world studies, we recover meaningful spatial and functional structure from aggregated data, while exposing the inherent limits of the setting: relative structure and causal dependencies are recoverable, but absolute local magnitudes may be systematically overestimated without fine-grained outcome supervision. 

Because the problem class is, to our knowledge, new in this explicit form, principled baselines are scarce and a complete identification theory does not yet exist. We see both as consequences of stating a new problem, and hope this paper motivates methods that supersede ours.

Although this paper primarily proposes a research framework, misuse in high-stakes domains such as epidemiology could cause harm if poorly identified local effects guided interventions or resource allocation; outputs should be accompanied by uncertainty estimates, sensitivity analyses, and domain expertise. Disaggregation itself may also raise privacy concerns, since inferred high-resolution patterns can reveal sensitive local information.

Several directions stand out. Theoretically, the most pressing need is a rigorous identifiability analysis: which combinations of aggregation function, contextual diversity, and structural restrictions guarantee recovery of the local mechanism, and which only up to monotone or scale transformations? Methodologically, principled regularization, hierarchical priors, or Bayesian formulations could control aggregation-induced bias and provide uncertainty quantification. Relaxing the shared-mechanism assumption, integrating causal discovery over contextual variables, incorporating instrumental-variable settings, and extending to spatiotemporal aggregation are natural next steps. We hope that framing \textsc{Clam} as a first concrete entry in a broader problem class makes these extensions easier to pursue.

\newpage

{\small
\bibliographystyle{unsrtnat}
\bibliography{example_paper}
}

\medskip


\appendix

\section{Identifiability}
\label{app:identifiability}
Identification in \textsc{Clam} depends on the target of interest and on the model assumptions. We distinguish causal identification under an assumed graph, statistical identification from aggregate observations, and finite-sample recovery by the optimizer. The last is an estimation problem rather than an identification question.

A complete theory for statistical identification from aggregates is beyond the scope of this work. Below, we characterize several tractable special cases and discuss the main remaining open problems.

\subsection{Identifiability of What}

First, \emph{causal identification} asks whether a quantity such as LOCATE can be expressed in terms of the observed distribution under the assumed causal graph. We treat the graph as given and do not consider causal discovery. This is separate from whether the required fine-grained mechanism can be recovered from aggregate observations.

Second, \emph{statistical identification} asks whether the local mechanism $f(\cdot)$ is uniquely determined by the population distribution of the aggregate data. We are generally interested in the function itself rather than the parameter vector $\boldsymbol{\theta}$, since different parameterizations may represent the same function.

Third, one can ask whether the high-resolution outcome map $\{y_{i,j}\}$ is identified. Even with a known mechanism, exact local outcomes may remain unknown because the individual noise realizations $\epsilon_{i,j}$ and possibly latent inputs $c'_{i,j}$ are unobserved. Conversely, aggregates can sometimes determine particular local outcomes without identifying the full mechanism.

\subsection{General Non-Identifiability}
Without restrictions, neither the mechanism nor the high-resolution outcome map is identified from aggregates.

Consider a noise-free setting similar to Equation~\eqref{eq:clam-sum} in which no region receives treatment. Take two regions $A$ and $B$, each with two subregions, with context values $(0,1)$ and $(0.5,0.5)$, respectively, and suppose both observed regional aggregates equal $1$.

If both the local mechanism $f_{\boldsymbol{\theta}}(\cdot)$ and the aggregation function $g_{\boldsymbol{\phi}}(\cdot)$ are unknown, infinitely many pairs are compatible with these observations. For example, $f_{\boldsymbol{\theta}}(0,c)=c$ together with sum aggregation yields subregional outcomes $(0,1)$ for region $A$ and $(0.5,0.5)$ for region $B$. Equally, the constant mechanism $f_{\boldsymbol{\theta}}(0,c)\equiv1$ together with mean aggregation yields $(1,1)$ for both regions. Both explain the observed aggregates while implying different high-resolution outcomes.

The ambiguity remains even when the aggregation function is known. Suppose $g_{\boldsymbol{\phi}}(\cdot)$ is a sum. Then $f_{\boldsymbol{\theta}}(0,c)=c$ is compatible with the data, but so is $f_{\boldsymbol{\theta}}(0,c)\equiv0.5$. The former yields $(0,1)$ in region $A$, whereas the latter yields $(0.5,0.5)$, while both sum to $1$.

\subsection{A Saturated View}
A useful limiting case is to think of $f_{\boldsymbol{\theta}}(\cdot)$ as a dictionary lookup: each distinct observed input $(t_{i,j},c_{i,j})$ is assigned an independent unknown value. With known sum aggregation and no noise, the aggregate observations define a linear system
\[
\mathbf{Y}=A\mathbf{v},
\]
where $\mathbf{v}$ contains the unknown values of $f_{\boldsymbol{\theta}}(\cdot)$ at the observed inputs. Identification is therefore a rank question: the local values are identified exactly when $A$ has full column rank. Otherwise, multiple local-value assignments produce the same aggregates.

A neural network departs from this saturated case by tying values at different inputs together through its parameterization and inductive bias. The linear model in Equation~\eqref{eq:clam-linear} imposes an even stronger restriction; there, identification analogously reduces to full rank of the corresponding low-dimensional design matrix.

\subsection{Aggregate Regression}
The saturated view above is deliberately pessimistic because it treats the value at every distinct input as unrelated. With a shared mechanism and sufficiently varied aggregate observations, much more can be identified. This is closely related to learning from aggregate observations as studied by \citet{zhang2020learning}.

The additive model in Equation~\eqref{eq:clam-sum} has a direct connection to their framework. Consider the special case of known sum aggregation, fixed $M$, additive homoscedastic Gaussian noise, no latent $c'$, and the sampling assumptions of \citet{zhang2020learning}. Since fixed-size sum and mean aggregation differ only by a constant scaling, the resulting aggregate MSE is equivalent to their regression-from-mean objective.

Let $[W_0]$ denote the equivalence class of all parameter values that are observationally equivalent to $W_0$, meaning that they induce the same population likelihood. Their Proposition~7 characterizes this class as
\begin{equation}
    [W_0]
    =
    \left\{
    W \in \mathcal{W}
    :
    f(X;W)=f(X;W_0)
    \ \text{a.e.}
    \right\}.
\end{equation}
Thus, the parameterization $W_0$ need not be unique, but all observationally equivalent parameters represent the same function almost everywhere under the distribution of $X$.

There is one subtlety in translating this result to our setting. Treatment is assigned at the regional level and is therefore shared by all subregions in a region. The result is therefore most naturally interpreted separately within each treatment arm. For a fixed $t\in\{0,1\}$, define
\[
    f_t(c)=f(t,c).
\]
If the subregional contexts within that treatment arm satisfy the sampling assumptions of \citet{zhang2020learning}, Proposition~7 identifies $f_t(\cdot)$ almost everywhere under the corresponding context distribution.

Consequently, recovering both $f_0(c)$ and $f_1(c)$ does not automatically identify LOCATE everywhere. The contrast
\[
    f_1(c)-f_0(c)
\]
is supported by this result only where the context distributions of the treated and untreated populations overlap. Outside this overlap, a flexible model relies on interpolation or extrapolation induced by its function class rather than on the identification result.

The independence assumptions are also restrictive for spatial data. The model of \citet{zhang2020learning} assumes conditional independence of the individual targets and treats the features within an aggregate as independent draws. Their proof of Proposition~7 uses this independence to factorize the cross terms in the population loss. Spatially structured covariate compositions need not satisfy these assumptions. \citet{zhang2020learning} themselves note that their independence assumption can fail when observations are collected group by group, explicitly naming spatially aggregated data as an example. Their theory therefore provides a useful base case and suitable language for functional equivalence, but it does not cover the full spatial setting considered here.

The counterexample above does not contradict this result: it uses only two fixed aggregate observations, which leave multiple mechanisms possible. Under the sampling assumptions of \citet{zhang2020learning}, many independently varying aggregates provide additional constraints that identify the shared function in the population limit.

\subsection{Open Problems}
The cases above provide a general counterexample and tractable results for restricted settings. Several important aspects of the general model remain open.

\paragraph{Spatio-Temporal Extensions and Latent Structure.}
Repeated spatio-temporal observations provide different information from additional independent aggregate groups. The same spatial units may recur over time with largely persistent spatial structure. A latent quantity $c'$ that is shared across observations therefore enters many aggregate constraints, whereas an idiosyncratic noise realization $\epsilon_{i,j}$ enters only one observation.

Repeated observations can consequently make persistent latent structure estimable while observation-specific noise remains unrecoverable. Exp.~3 illustrates this empirically: the same latent vegetation field contributes to regional outcomes over many time points, allowing constraints on it to accumulate. We do not claim a general identification theorem for this setting; recovery depends on temporal variation, the functional form, and how the latent variable enters the mechanism.

\paragraph{Unknown Aggregation.}
If $g_{\boldsymbol{\phi}}(\cdot)$ is unknown and learned jointly with $f_{\boldsymbol{\theta}}(\cdot)$, different combinations of the local mechanism and aggregation function may induce the same observed regional outcomes, as illustrated by the first counterexample. Identification therefore requires restrictions on at least one of these functions, sufficient variation to distinguish them, or prior knowledge of the aggregation process.

This is one reason why known sum or mean aggregation is substantially easier than the unrestricted model in Equation~\eqref{eq:clam-general}.

\paragraph{Nonlinear Aggregation and Noise.}
Linear aggregation has the convenient property that, with additive zero-mean noise, aggregation commutes with expectation. For a nonlinear aggregation function this generally fails:
\begin{equation}
g_{\boldsymbol{\phi}}
\left(
\left\{
\mathbb{E}[y_{i,j}\mid T,C]
\right\}_{j}
\right)
\neq
\mathbb{E}
\left[
g_{\boldsymbol{\phi}}
\left(
\{y_{i,j}\}_{j}
\right)
\mid T,C
\right].
\end{equation}

Applying a nonlinear aggregator directly to noise-free conditional-mean predictions therefore does not generally target the conditional mean of the observed aggregate. A statistically correct objective must propagate the noise through the aggregation, for example through the induced likelihood or a Monte Carlo approximation. This removes the objective mismatch but does not by itself establish identification. If $f_{\boldsymbol{\theta}}(\cdot)$, $g_{\boldsymbol{\phi}}(\cdot)$, or the noise distribution are simultaneously unknown, further observational equivalences may remain.

\paragraph{Causal Structure.}
The statistical arguments above concern recovery of a function from aggregate data and do not, by themselves, give that function a causal interpretation. We do not attempt to discover the causal graph. Instead, we assume a graph and ask whether, under that graph, the recovered function corresponds to the desired interventional quantity.

For the baseline motif, where $C$ is an observed effect modifier and treatment is unconfounded conditional on the relevant variables, $f(t,c)$ can represent the conditional interventional mean required for LOCATE. If $C$ is a confounder, the treatment-assignment mechanism must be handled appropriately. If $C$ is a mediator or collider, conditioning on it has a different causal meaning and may not correspond to the desired treatment effect. Thus, changing the assumed causal structure need not change the aggregate regression problem mathematically, but it changes which recovered functions correspond to identified causal estimands.

\paragraph{Identification Versus Optimization.}Even when the population objective identifies the desired function, this does not imply that a finite-sample non-convex optimizer will recover it. Neural-network parameterization, jointly inferred latent variables, and learned aggregation functions may create multiple local optima or flat regions of the objective. Conversely, failure of one optimizer to recover the ground truth does not itself imply non-identifiability.

We therefore treat optimization stability as an empirical question separate from identification. In Exp.~1, repeated random initializations produce similar LOCATE estimates when the aggregate loss is low, but this is evidence about the estimator in that experiment, not proof that the population problem is identified.

\subsection{Practical Diagnostics}
No finite-sample diagnostic can prove structural identification. Nevertheless, several checks can reveal that the aggregate observations do not sufficiently constrain the fine-grained quantity of interest.

We recommend comparing recovered LOCATE estimates across random restarts, architectures, regularization choices, and plausible aggregation or noise specifications. Large disagreement is evidence of an underdetermined problem. Agreement is reassuring but does not establish identification, since the same inductive bias may repeatedly select the same solution.

External validation against higher-resolution observations should be used whenever available. Sensitivity analyses under plausible violations of the causal assumptions provide complementary information; our hidden-confounding ablation is one example. Bootstrap intervals quantify sampling variability but should not be interpreted as a test of structural identification.

\paragraph{Summary of Claims.}
The discussion can be summarized in three levels.

\begin{enumerate}
    \item For the linear model in Equation~\eqref{eq:clam-linear}, identification reduces to a rank condition on the aggregate design matrix.

    \item For the additive model in Equation~\eqref{eq:clam-sum}, the i.i.d.\ mean-aggregation special case connects directly to the functional consistency result of \citet{zhang2020learning}. Within each treatment arm, their result identifies the conditional-mean function almost everywhere under the corresponding context distribution. Identification of LOCATE additionally requires sufficient overlap between the treatment arms.

    \item For the general model in Equation~\eqref{eq:clam-general}, including dependent spatial designs, structured latent variables, unknown or nonlinear aggregation, and extrapolation outside the observed treatment--context support, we do not claim a general identification theorem. Recovery in these regimes is studied empirically and remains an important theoretical problem.
\end{enumerate}

Causal identification under the assumed graph is separate from all three statements: even a statistically identified fine-grained mechanism has the desired causal interpretation only under the assumptions encoded by that graph.

\section{Causal Deabstraction Framework}\label{app:CausalDeabstraction}
In this section, we situate our work within the broader causal inference literature. Specifically, we frame this as a novel instance of \emph{causal deabstraction} \cite{beckers2019abstracting,zennaro2022abstraction}. Our methodology is informed by the framework of SCMs and is guided by the principle of ICM.

\subsection{Structural Causal Framework}
The structural causal model introduced in Section~\ref{sec:scmmain} can be formalized in generative form as follows. Each spatial unit (region) $i \in \{1, \dots, N\}$ is composed of $M$ subregions indexed by $j \in \{1, \dots, M\}$. For each subregion $(i,j)$, we posit the following SCM:
\begin{align*}
c_{i,j} &\ \text{observed (exogenous)} \\
\epsilon_{i,j} &\sim P_\epsilon \\
t_{i,j} &= T_i \in \{0,1\} \\
y_{i,j} &= f_{\boldsymbol{\theta}}(t_{i,j}, c_{i,j}, \epsilon_{i,j}) \\
\widehat{Y}_i &= g_{\boldsymbol{\phi}}(\{y_{i,j}\}_{j=1}^M)
\end{align*}
Here, $c_{i,j}$ denotes a high-resolution contextual covariate for subregion $j$ within region $i$, which is observed and exogenous, meaning it is not caused by $T$ or $Y$, and $\epsilon_{i,j}$ represents exogenous noise drawn from $P_\epsilon$. The binary intervention $t_{i,j} = T_i \in \{0,1\}$ is assigned at the regional level and shared across all subregions. The function $f_{\boldsymbol{\theta}}(\cdot)$ encodes the local causal mechanism, shared across all subregions, that maps the intervention, context, and noise to the subregional outcome $y_{i,j}$. The aggregation function $g_{\boldsymbol{\phi}}(\cdot)$ maps the collection of subregional outcomes $\{y_{i,j}\}_{j=1}^M$ to the regional outcome $\widehat{Y}_i$.

In the main section, we considered simplified cases, such as sum aggregation, additive noise, and the absence of hidden confounders, to provide intuition. The formulation here generalizes that view by expressing the structural causal model in explicit stochastic form, specifying how interventions, contextual variables, and noise are generated, while making clear that both $f_{\boldsymbol{\theta}}(\cdot)$ and $g_{\boldsymbol{\phi}}(\cdot)$ are shared across regions.

\subsection{Invariant Causal Mechanism Assumption}
A central assumption in \textsc{Clam} is that the causal function $f_{\boldsymbol{\theta}}(\cdot)$ is \emph{invariant} across all subregions and regions. That is, while the covariates $\{c_{i,j}\}$ may vary across regions due to differences in local composition, the functional form $f_{\boldsymbol{\theta}}(t, c)$ does not change. This assumption reflects the principle of \emph{invariant causal mechanisms} \cite{scholkopf2021toward,peters2016causal,heinze2018invariant}, which posits that causal relations are stable across environments unless directly intervened upon.

We treat each region $i$ as an \emph{environment} defined by its observed covariate composition ${c_{i,j}}*{j=1}^M$, but governed by a shared mechanism $f*{\boldsymbol{\theta}}(\cdot)$. The variability in covariate composition across regions provides information for estimating $f_{\boldsymbol{\theta}}(\cdot)$ from the observed aggregated outcomes $\widehat{Y}_i$.

\subsection{Deabstraction via Invariance}
We propose to view the task of estimating $f_{\boldsymbol{\theta}}(\cdot)$ from aggregated outcomes as an instance of \emph{causal deabstraction}, that is, inferring a latent, high-resolution causal model that is consistent with a lower-resolution aggregated model. This perspective is dual to recent work on causal abstraction \cite{beckers2019abstracting,zennaro2022abstraction}, which studies when a coarse model preserves the counterfactual semantics of a fine-grained one. Here, we reverse the direction: we seek to recover a fine-grained causal mechanism that is \emph{compatible} with the observed coarse-level effects.

This is possible due to two key properties:
\begin{enumerate}
    \item The aggregation function $g_{\boldsymbol{\phi}}(\cdot)$ is consistent and applied uniformly across regions.
    \item The observed covariate compositions $\{c_{i,j}\}_{j=1}^M$ differ across regions, which induces identifiable variation in the aggregates $\widehat{Y}_i$ under the shared mechanism $f_{\boldsymbol{\theta}}(\cdot)$.
\end{enumerate}
Given sufficient diversity and the assumption of causal invariance, the aggregated outputs provide a supervisory signal to recover the latent causal function (although we do not provide a general correctness guarantee).

\subsection{Causal Semantics via Do-Calculus}
\label{sec:docalc}
Under the assumption of no hidden confounding, the causal effect of the intervention $T_i$ on a subregional outcome $y_{i,j}$ is identified via the backdoor criterion:
\begin{equation*}
    \mathbb{E}[y_{i,j} \mid do(T_i = t),\, c_{i,j}] = \int f_{\boldsymbol{\theta}}(t, c_{i,j}, \epsilon_{i,j}) \, p(\epsilon_{i,j}) \, d\epsilon_{i,j}
\end{equation*}
In Exps~1--4, we assume $T_i \perp \!\!\! \perp \epsilon_{i,j} \mid c_{i,j}$, and that $T_i$ is either randomized or independent of $C_{i,j}$. Under this setting, the expectation can be approximated via observational data. However, because only the aggregate outcome $\widehat{Y}_i$ is observed, we must rely on the consistency constraint:
\begin{equation*}
    \widehat{Y}_i \approx \sum_{j=1}^M f_{\boldsymbol{\theta}}(T_i, c_{i,j})
\end{equation*}
The training procedure thus finds the function $f_{\boldsymbol{\theta}}(\cdot)$ that jointly explains all observed aggregates under this constraint, while enforcing that $f_{\boldsymbol{\theta}}(\cdot)$ is invariant across regions. In this sense, the estimation of $f_{\boldsymbol{\theta}}(\cdot)$ becomes a constraint satisfaction problem, where causal invariance and aggregation consistency define the feasible solution space.
In Exp~5, this conditional independence assumption is deliberately violated by construction, and valid estimation requires explicitly modeling the treatment allocation mechanism.

\subsection{Implications}
The assumption of an invariant subregional mechanism plays a central role in enabling causal inference in our disaggregated setting. By treating the coarse observations as aggregates of fine-grained causal effects and leveraging variation in covariate compositions, \textsc{Clam} performs inference not merely on statistical associations but on causal mechanisms that are meaningful and robust to changes in population structure. All in all, we contribute a formal and algorithmic instantiation of \emph{causal deabstraction} from low-resolution outcomes and high-resolution covariates.

\section{Extended Related Work}\label{app:relatedwork}

Our work builds on and extends three main strands of research: spatial disaggregation, causal inference under aggregation, and causal abstraction.

\paragraph{Statistical Downscaling and Disaggregation.} Statistical downscaling methods aim to infer high-resolution estimates from coarse data, using spatial interpolation, small area estimation \cite{fay1979estimates}, or learning-based techniques. These have been widely applied in climate and environmental sciences \cite{maraun2019statistical, peng2017review, gotway2002combining}, including recent work on probabilistic downscaling using Gaussian processes \cite{chau2021deconditional, law2018variational}. However, such methods typically model associational patterns and do not support interventional or counterfactual reasoning. 

To address this, some recent approaches integrate causal reasoning into downscaling pipelines, e.g., by identifying causally relevant predictors \cite{vazquez2022effectiveness, dutta2020identification}. Yet, these still rely on observable variables and assume access to relatively fine-grained information. In contrast, our approach estimates high-resolution \textit{causal effects} using only aggregated outcome data and high-resolution covariates. 

The challenges of reasoning with aggregate data are well-documented. Classical results on the ecological fallacy show that aggregated statistics may obscure or even invert causal relationships \cite{robinson1950ecological, freedman1999ecological, king1997solution}. Earlier work on aggregation and disaggregation in optimization similarly highlights the complexity introduced when latent heterogeneity is present but unobserved \cite{rogers1991aggregation}. \textsc{Clam} explicitly addresses these challenges by modeling the aggregation process and learning disentangled causal effects at the subregional level.

\paragraph{Causal Deabstraction.}
Structurally, our approach is inspired by recent work on causal abstraction, which studies mappings between causal models at different levels of granularity \cite{beckers2019abstracting, beckers2020approximate, zennaro2022abstraction}. These works formalize when such mappings preserve counterfactual or interventional semantics, but do not address learning disaggregated effects from data. Finally, our work relates to spatiotemporal causal discovery \cite{runge2019inferring}, but differs by operating in a setting where temporal data is limited and only aggregate observations are available.

\paragraph{Graph Neural Networks.}
Conceptually, our architecture is related to message-passing graph neural networks \cite{gilmer2017neural}. In a GNN, pairwise messages are computed and aggregated within a single update step. Similarly, our model learns an update function $f_{\boldsymbol{\theta}}(\cdot)$ that operates on ordered inputs, and optionally an aggregation function $g_{\boldsymbol{\phi}}(\cdot)$ that is permutation-invariant, akin to a Deep Sets model \cite{zaheer2017deep}. As in GNNs, we share the parameters of the structural function across nodes or regions.

\paragraph{Invariant Causal Mechanisms.}
Our work also relates to the principle of \emph{invariant causal mechanisms}, which posits that the functional relationships between variables remain stable across different environments or interventions \cite{peters2016causal, scholkopf2021toward}. This idea has been used for causal discovery \cite{heinze2018invariant} and domain generalisation \cite{rojas2018invariant}, and is particularly relevant for spatial disaggregation where intervention effects may vary locally but share stable dependencies on contextual covariates. \textsc{Clam} incorporates this principle by enforcing that the learned local causal mechanism $f_{\boldsymbol{\theta}}(\cdot)$ is shared across all subregions while allowing spatial heterogeneity to emerge through contextual variables.

\paragraph{Hierarchical Bayesian Model.}
The design also parallels hierarchical Bayesian models \cite{feller2015hierarchical,dias2013hierarchical}, where group-specific effects are drawn from a shared prior distribution and information is partially pooled across groups. In \textsc{Clam}, the local mechanism $f_{\boldsymbol{\theta}}(\cdot)$ plays the role of the shared prior, and contextual covariates capture structured variation across subregions, allowing data-sparse subregions to benefit from patterns learned in others.

\paragraph{Spatiotemporal Causality.}
Spatiotemporal causal inference has received growing attention as researchers seek to understand how interventions and their effects evolve across both space and time. A recent work presents a causal framework for studying the interplay between conflict and forest loss in Colombia, emphasizing the need to model spatial and temporal dependencies jointly \cite{christiansen2022toward}. Extending this perspective, methods proposed in~\cite{zhou2024estimating} estimate heterogeneous treatment effects in settings where economic assistance moderates the impact of airstrikes on insurgent violence, while approaches introduced in~\cite{mukaigawara2025spatiotemporal} accommodate arbitrary spill-over and carryover effects across space and time. Network-based settings with temporal dynamics have also been explored. In~\cite{song2024bipartite} the authors study bipartite causal inference with interference in time series data, and in~\cite{balkus2024causal} methods are developed to analyse the causal effects of modified treatment policies under network interference. Recent methodological innovations include GST-UNet~\cite{oprescu2026gst}, a deep learning architecture for spatiotemporal causal inference under time-varying confounding, and consistency-guided aggregation for causal discovery in multivariate spatiotemporal data introduced in~\cite{ninad2025causal}. Together, these works highlight the diversity of approaches to spatiotemporal causality, spanning structural models, interference-aware estimation, and machine learning-driven inference.


\newpage
\section{Details of Main Experiments}
\label{app:expdetails}

\subsection{Exp.~1: Political Campaigning}
\label{exp:1details}

\paragraph{Context.}
We consider a simple political campaigning setting in which a candidate either spends campaign resources in a region or does not. The treatment is binary at the regional level and is inherited by all subregions. The outcome is the candidate's relative improvement, but it is observed only as a regional aggregate. The effect of campaigning depends on subregional context, here interpreted as standardized wealth. Thus, two regions can have similar aggregate context and similar aggregate outcomes while still containing different subregional causal effects.

\begin{figure*}[t]
    \centering
    \includegraphics[width=0.9\linewidth]{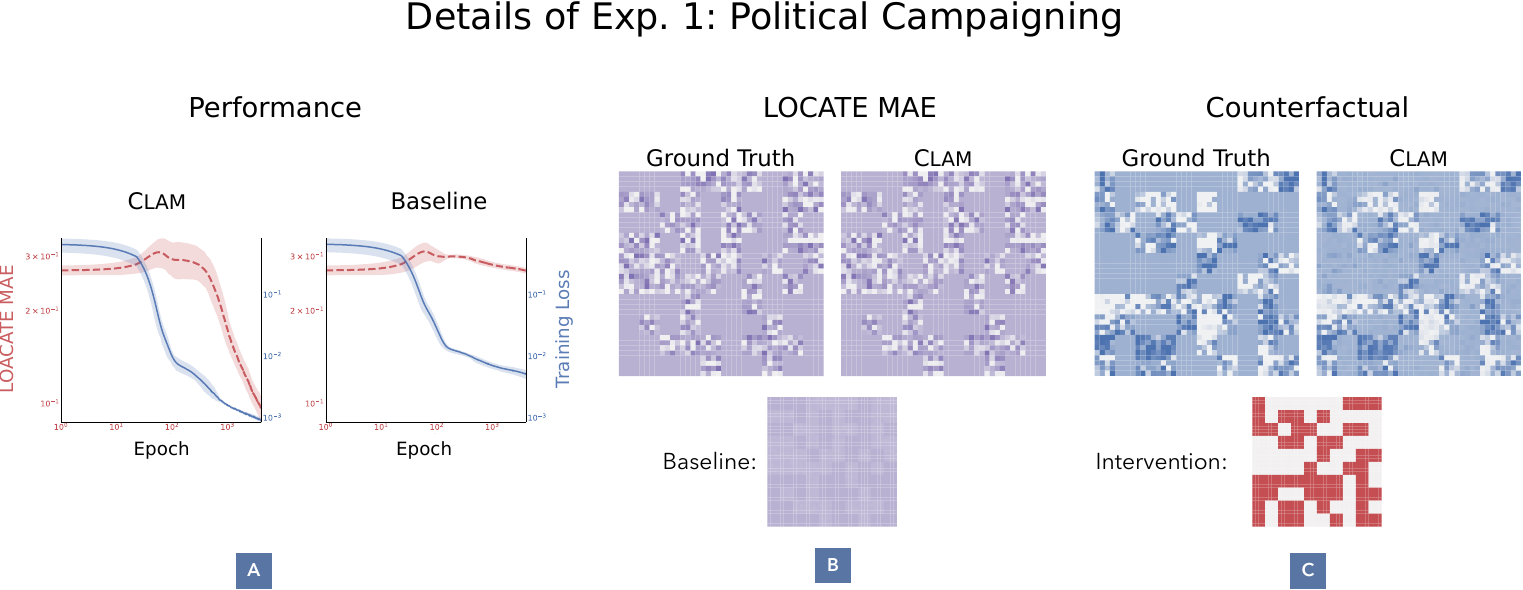}
    \caption{Evaluation of the learned outcomes, causal effect estimates, counterfactual predictions, and training behavior.}
    \label{fig:results_exp1}
\end{figure*}

\paragraph{Setup.}
We use \(N=100\) regions arranged on a \(10 \times 10\) grid. Each region contains \(M=16\) subregions arranged on a \(4 \times 4\) grid. The treatment matrix is \(T \in \{0,1\}^{N \times M}\). Exactly half of the regions are treated, and within a treated region all subregions have treatment value \(1\); within a control region all subregions have treatment value \(0\).

The context matrix \(C \in \mathbb{R}^{N \times M}\) contains one continuous contextual value per subregion. We sample the entries from a standard Gaussian distribution and then subtract the mean within each region, so that each row of \(C\) has mean zero. This makes the regional average context uninformative by construction. We then randomly swap entries across the matrix to introduce additional variation while preserving the overall structure. The observed outcome is generated at the subregional level and then aggregated to the region level by averaging.

The ground-truth outcome function is
\[
f(t,c) =
\begin{cases}
1, & t = 0, \\
\operatorname{interp}(c), & t = 1,
\end{cases}
\]
where \(\operatorname{interp}(c)\) is the piecewise linear interpolation through the anchor points $\{ (-2,2),(0,1),(1,0) \}$.

Thus, untreated subregions have outcome \(1\), while treated subregions have an outcome that depends nonlinearly on the local context \(c\). The noisy subregional outcome is
\[
y_{i,j} = f(t_{i,j}, c_{i,j}) + \epsilon_{i,j},
\qquad
\epsilon_{i,j} \sim \mathcal{N}(0, 0.1^2),
\]
and the observed regional outcome is
\[
\widehat{Y}_i = \frac{1}{M}\sum_{j=1}^{M} y_{i,j}.
\]
The ground-truth local causal effect, which we use only for evaluation, is
\[
E_{i,j} = f(t_{i,j}, c_{i,j}) - f(0, c_{i,j}).
\]

\paragraph{Training.}
We train a small neural network \(f_{\boldsymbol{\theta}}(t,c)\) with two inputs, treatment and context, and one scalar output. The model has two hidden layers with \(32\) hidden units each, ReLU activations, dropout with probability \(0.05\), and is trained with AdamW using learning rate \(10^{-3}\) and weight decay \(10^{-3}\) for \(4000\) epochs. Importantly, \textsc{Clam} is trained only from the regional aggregate signal. For each region, we average the model's subregional predictions and minimize
\[
\mathcal{L}_{\textsc{Clam}}
=
\frac{1}{N}
\sum_{i=1}^{N}
\left(
\frac{1}{M}\sum_{j=1}^{M} f_{\boldsymbol{\theta}}(t_{i,j}, c_{i,j})
-
\widehat{Y}_i
\right)^2 .
\]

\paragraph{Baseline.}
We compare against a uniform disaggregation baseline. Since the true subregional outcomes are not observed, this baseline assigns the same regional outcome to every subregion,
\[
\widetilde{y}_{i,j} = \widehat{Y}_i .
\]
It then trains the same neural network architecture using an element-wise loss,
\[
\mathcal{L}_{\text{base}}
=
\frac{1}{NM}
\sum_{i=1}^{N}
\sum_{j=1}^{M}
\left(
f_{\boldsymbol{\theta}}(t_{i,j}, c_{i,j})
-
\widetilde{y}_{i,j}
\right)^2 .
\]
This gives the baseline a dense subregional loss signal, but the signal is based on a uniformity assumption that removes the within-region heterogeneity needed to recover local causal effects.

\paragraph{Evaluation.}
We evaluate whether the learned model can recover the fine-grained causal structure from aggregate observations (Figure~\ref{fig:results_exp1}). We report the training loss and LOCATE MAE (cf.\ Eq.~\eqref{eq:simplemaineq}) over training (Figure~\ref{fig:results_exp1}a).  Confidence intervals (95\%) are bootstrapped and based on 20 runs.
As expected, both decrease for \textsc{Clam}, while the baseline cannot learn the causal mechanism. We also report the full LOCATE matrix, comparing the ground truth with the one inferred by \textsc{Clam}, in Figure~\ref{fig:results_exp1}b.
The recovered causal effect is almost indistinguishable from the ground truth, while the baseline, as expected, cannot reconstruct the underlying disaggregation.
Lastly, we construct a counterfactual treatment matrix (Figure~\ref{fig:results_exp1}c) by inverting the observed treatment assignment, \(T^{\mathrm{cf}} = 1 - T\), and compare the ground-truth counterfactual outcome \(f(T^{\mathrm{cf}}, C)\) with the neural-network estimate \(f_{\boldsymbol{\theta}}(T^{\mathrm{cf}}, C)\). This tests whether the learned mechanism supports coherent counterfactual prediction beyond the observed treatment assignment.

\begin{figure}[t]
    \centering
    \includegraphics[width=0.4\linewidth]{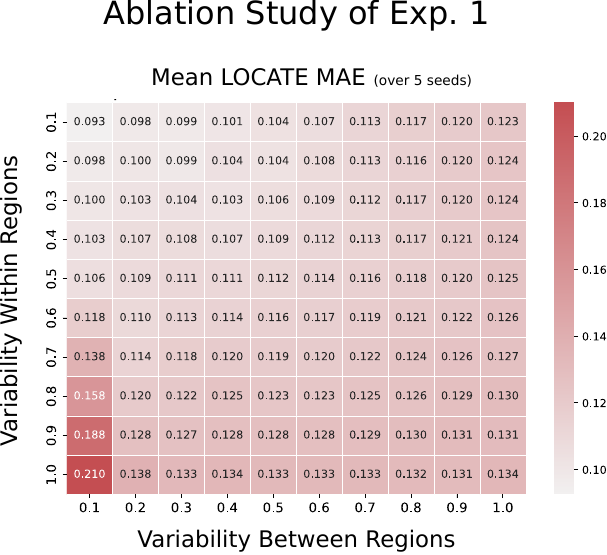}
    \caption{We evaluate the effect of varying the variance of regional context means and, for each fixed context mean, the inter-regional variance. Results are averaged over 5 random seeds for each parameter combination, and uncertainty bounds represent the corresponding variability.}
    \label{fig:ablation}
\end{figure}

\paragraph{Ablation.}
To test when this recovery is possible, we repeat the experiment while varying two properties of the context matrix. The first is the standard deviation of the regional context means, denoted \(\sigma_{\text{reg}}\). The second is the standard deviation of context values within each region, denoted \(\sigma_{\text{subreg}}\). For each pair \((\sigma_{\text{reg}}, \sigma_{\text{subreg}})\), we run the experiment over multiple random seeds and average the final LOCATE MAE. This ablation shows that high variability within regions is associated with a smaller error.

\subsubsection{Ablation of Exp.~1: Political Campaigning}
Our setting targets low-variability aggregate regimes, where regional outcomes are nearly constant and the inverse problem is underdetermined: many fine-grained causal explanations can yield indistinguishable regional observations. The key idea is that high-dimensional subregional covariates can restore information by exposing the shared causal mechanism \(f_{\boldsymbol{\theta}}(\cdot)\) to diverse contextual compositions within each aggregate unit. Exp.~1 and the corresponding ablation support this mechanism: as subregional heterogeneity increases, recovery of the true causal effect improves, even when aggregate outcomes remain similarly uninformative. By contrast, varying regional-level heterogeneity has a weaker effect, indicating that the gains do not come primarily from more variable aggregate outcomes, but from richer within-region contextual structure. See Figure \ref{fig:ablation} for details. Thus, causal deabstraction becomes possible in otherwise weakly identified settings when subregional covariate variation, together with an invariant causal mechanism, provides enough compositional diversity to distinguish the underlying fine-grained causal function.

\subsection{Exp.~2: Public School Funding vs Improvement in Outcomes}
\label{exp:2details}

\begin{figure*}[t]
    \centering
    \includegraphics[width=0.9\linewidth]{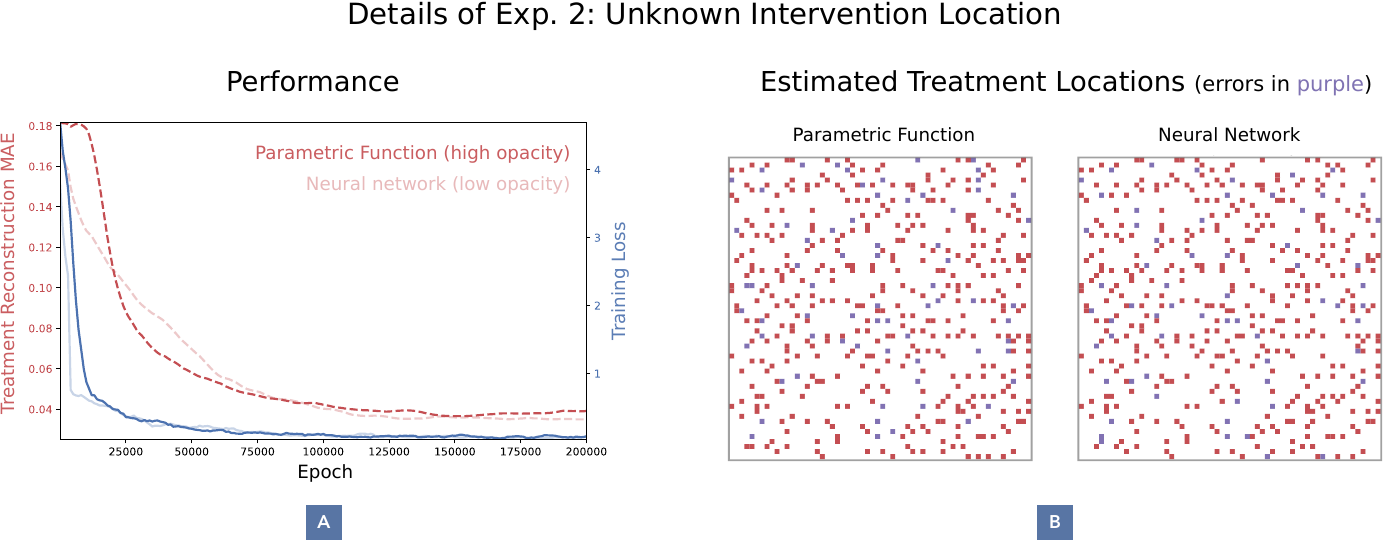}
    \caption{Performance of the training loss and MSE of the local-effect estimation, alongside ground-truth causal effects, estimated causal effects, and the corresponding estimation errors across subregions.}
    \label{fig:exp2details}
\end{figure*}

\paragraph{Context.}
We examine how public spending on schools affects educational outcomes when the exact locations of the expenditures are \emph{unknown}. Each administrative region contains multiple subregions, interpreted here as school districts, but only one district in a treated region receives a funding intervention. At the regional level, we only observe an aggregated treatment indicator and the aggregated outcome, not the specific district that received the intervention. The goal is to infer the high-resolution intervention locations from aggregated data.

The subregional context consists of high-resolution demographic and economic data, represented here by a normally distributed \emph{wealth score}. We hypothesize that the effect of school funding depends on this wealth score through a global shift and scale.

\paragraph{Setup.}
We assume $N = 30 \times 30 = 900$ regions, each with $M = 2 \times 2 = 4$ subregions.

We simulate:
\begin{itemize}
    \item A high-resolution intervention matrix $T \in \mathbb{R}^{N \times M}$, where $t_{i,j}$ indicates whether subregion $j$ of region $i$ receives the funding intervention.
    \item A regional intervention vector $\widehat{T} \in \mathbb{R}^N$, where
    $
        \widehat{T}_i = \sum_{j=1}^M t_{i,j}
    $
    is the absolute intervention intensity in region $i$.
    \item A high-resolution context matrix $C \in \mathbb{R}^{N \times M}$ encoding the socioeconomic status of each subregion, where smaller $c_{i,j}$ values represent lower socioeconomic status and larger $c_{i,j}$ values represent higher socioeconomic status.
    \item A high-resolution noise matrix $\epsilon \in \mathbb{R}^{N \times M}$ with i.i.d.\ entries $\epsilon_{i,j} \sim \mathcal{N}(0, \sigma^2)$, where $\sigma^2 = 0.005$.
\end{itemize}

The intervention matrix $T$ is generated by:
\begin{enumerate}
    \item Initializing all entries to zero.
    \item Randomly selecting $50\%$ of the regions for treatment.
    \item For each treated region $i$, choosing exactly one subregion $j \in \{1,\dots,M\}$ uniformly at random and assigning $t_{i,j}=1$.
\end{enumerate}

The context matrix $C$ is sampled i.i.d.\ from a standard normal distribution:
\[
    c_{i,j} \sim \mathcal{N}(0,1).
\]

The outcome variable is given by the ground-truth functional relationship
\[
    y_{i,j}
    =
    f_{\boldsymbol{\theta}}\!\left(t_{i,j}, c_{i,j}\right)
    + \epsilon_{i,j}
    =
    \left(\theta_{\mathrm{shift}} - c_{i,j}\right)
    \cdot \theta_{\mathrm{scale}}
    \cdot t_{i,j}
    + \epsilon_{i,j},
\]
with ground-truth parameters
\[
    \theta_{\mathrm{shift}} = 4.2,
    \qquad
    \theta_{\mathrm{scale}} = 1.23.
\]

The observed regional outcome is aggregated by summation:
\[
    \widehat{Y}_i = \sum_{j=1}^M y_{i,j}.
\]

As in Exp.~1, we assume no hidden confounding, no time-series effects, treatment assignment independent of context, and aggregation-based regional observations.

\paragraph{Training.}
We train a model to estimate both the latent subregional intervention assignments and the parameters of the causal effect function.

The model has:
\begin{itemize}
    \item $N \times M = 900 \times 4$ latent treatment logits, one vector of length $M$ for each region.
    \item Two scalar parameters $\{\theta_{\mathrm{shift}}, \theta_{\mathrm{scale}}\}$ specifying the parametric form of $f_{\boldsymbol{\theta}}(\cdot)$.
\end{itemize}

The learned causal effect function is parametrized as
\[
    f_{\boldsymbol{\theta}}(t_{i,j}, c_{i,j})
    =
    \left( \theta_{\mathrm{shift}} - c_{i,j} \right)
    \cdot \theta_{\mathrm{scale}}
    \cdot t_{i,j}.
\]
To ensure a positive scale parameter, $\theta_{\mathrm{scale}}$ is represented through a softplus transformation.

An auxiliary preprocessing function is applied to the latent treatment logits using the Gumbel-Softmax relaxation. This produces a soft row-wise assignment over the $M$ subregions. The resulting assignment is processed so that untreated rows remain all-zero and treated rows have approximately one active subregion. During training, the Gumbel-Softmax temperature is exponentially annealed from $2.0$ to $0.1$.

In the forward pass:
\begin{enumerate}
    \item Sample a mini-batch of region indices.
    \item Process the latent treatment logits using the Gumbel-Softmax relaxation to obtain an estimate of the high-resolution treatment matrix $T$ for the mini-batch.
    \item Compute predicted subregional outcomes:
    \[
        \mu_{i,j}
        =
        f_{\boldsymbol{\theta}}\!\left(\widehat{t}_{i,j}, c_{i,j}\right).
    \]
    \item Aggregate to regional predictions by summation:
    \[
        \mu_i
        =
        \sum_{j=1}^M \mu_{i,j}.
    \]
    \item Compare the regional prediction with the observed regional outcome $\widehat{Y}_i$ using the MSE loss.
\end{enumerate}

We optimize all parameters jointly using AdamW with learning rate $10^{-4}$. Training is performed for $200{,}000$ epochs with a fixed random seed.

\paragraph{Neural Network.}
Additionally, we train a neural-network variant following the same treatment-assignment procedure. In this version, the causal effect function $f_{\boldsymbol{\theta}}(\cdot)$ is implemented as a feed-forward neural network taking $(t_{i,j}, c_{i,j})$ as input.

The network has three hidden layers with hidden dimension $16$ and $\tanh$ activations:
\[
    (t_{i,j}, c_{i,j})
    \mapsto
    \mathrm{MLP}_{\boldsymbol{\theta}}(t_{i,j}, c_{i,j}).
\]
To stabilize the local-effect interpretation, the model uses the centered form
\[
    f_{\boldsymbol{\theta}}(t_{i,j}, c_{i,j})
    =
    \mathrm{MLP}_{\boldsymbol{\theta}}(t_{i,j}, c_{i,j})
    -
    \mathrm{MLP}_{\boldsymbol{\theta}}(0, c_{i,j}),
\]
which enforces $f_{\boldsymbol{\theta}}(0,c_{i,j}) \approx 0$. The latent treatment indicators are processed in the same way as in the parametric model, using Gumbel-Softmax with temperature annealing from $2.0$ to $0.1$.

The neural-network model is also trained for $200{,}000$ epochs using AdamW with learning rate $10^{-4}$ and mini-batches containing $20\%$ of the regions.

\paragraph{Results.}
Figure~\ref{fig:exp2details}a shows the evolution of the mini-batch training loss and the quality of the treatment-location reconstruction (measured as MAE). Both the neural network and the parametric function recover the locations reasonably well. We however find that this depends on the noise level and on proper temperature annealing. The final reconstructed locations are shown in Figure~\ref{fig:exp2details}b, where incorrect predictions are shown in purple.

\subsection{Exp.~3: Spatiotemporal Effects of Heat Waves on School Performance}
\label{exp:3details}

\begin{figure*}[t]
    \centering
    \includegraphics[width=0.9\linewidth]{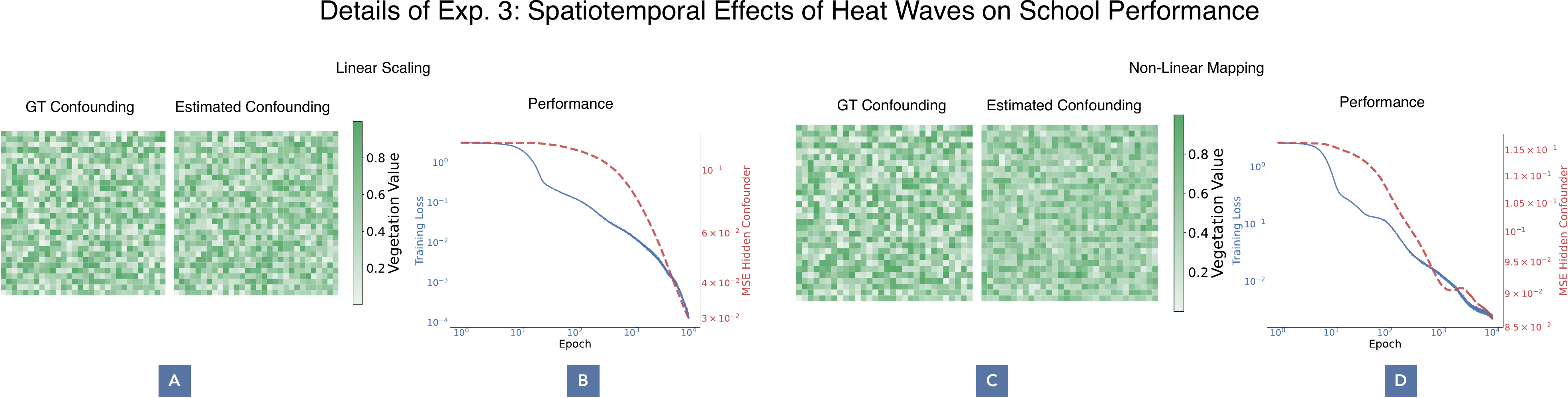}
    \caption{Ground-truth vegetation index, the corresponding estimated latent effect-modifier fields, and joint plots of training loss and vegetation MSE. Results are shown for two settings: linear scaling (left) and MLP learning (right).}
    \label{fig:results_exp3}
\end{figure*}

\paragraph{Context.}  
In this experiment, we examine the impact of extreme heat events (\say{heat waves}) on educational outcomes across both space and time. We model a spatial grid of $N = 100$ regions, each containing $M = 9$ subregions, observed over $W = 48$ consecutive months. The intervention represents the occurrence of a heat wave in a given subregion and month. Heat waves have a detrimental effect on school performance, but the strength of this effect depends on the educational background of parents in the subregion and on an unobserved environmental factor, vegetation coverage, which mitigates the impact of heat.

The observed context for each subregion is a categorical indicator of parents’ education level, taking one of three values (\emph{low} = 1, \emph{medium} = 2, \emph{high} = 3), which evolves slowly over time to reflect gradual demographic changes. The unobserved context is a fixed vegetation index between $0$ and $1$ for each subregion, representing the proportion of green cover and its protective effect against heat. The intervention process is binary at the subregional level, with most regions remaining unaffected in a given month, but some experiencing heat waves simultaneously in all their subregions, corresponding to large-scale weather patterns.

The outcome, measured as a monthly change in school performance for each subregion, is driven by the interaction between the intervention, parents’ education level, and the vegetation coverage. Subregions with lower parental education are more strongly affected, while vegetation dampens the negative effect. The challenge in this experiment is to recover both the mapping from context, intervention, and unobserved effect modifier to outcomes, and the vegetation values themselves, given only aggregated regional-level outcomes over time.

\paragraph{Setup.}  
We model $N = 100$ regions, each with $M = 9$ subregions, observed over $W = 48$ discrete time points (months).  

For each month $w \in \{1, \dots, W\}$, we define:
\begin{itemize}
    \item A high‐resolution binary treatment matrix $T^{(w)} \in \{0,1\}^{N \times M}$, where $t_{i,j}^{(w)} = 1$ indicates that subregion $j$ of region $i$ is experiencing a heat wave in month $w$. In our data generation, each row of $T^{(w)}$ is either all zeros (probability $0.7$) or all ones (probability $0.3$), corresponding to large‐scale heat events affecting entire regions.
    \item An observed context matrix $C^{(w)} \in \{1,2,3\}^{N \times M}$ encoding the categorical education level of parents in each subregion. At $w=1$, these values are sampled uniformly. For $w > 1$, $C^{(w)}$ is obtained from $C^{(w-1)}$ by flipping each entry to a random category with probability $0.05$, capturing slow demographic change.
    \item An unobserved covariate context matrix $U \in [0,1]^{N \times M}$ containing fixed vegetation index values for each subregion, drawn uniformly from $[0,1]$ and constant across all months.
    \item A noise matrix $\epsilon^{(w)} \in \mathbb{R}^{N \times M}$ with i.i.d.\ entries $\epsilon_{i,j}^{(w)} \sim \mathcal{N}(0, \sigma^2)$, with $\sigma^2 = 0.02$.
\end{itemize}

The subregional outcome is generated as:
\[
y_{i,j}^{(w)} = f_{\boldsymbol{\theta}}\!\left(t_{i,j}^{(w)}, c_{i,j}^{(w)}, u_{i,j}\right) + \epsilon_{i,j}^{(w)},
\]
where $c_{i,j}^{(w)}$ is the parents’ education category, $u_{i,j}$ is the vegetation index, and:
\[
f_{\boldsymbol{\theta}}\!\left(t_{i,j}^{(w)}, c_{i,j}^{(w)}, u_{i,j}\right) =
\begin{cases}
0 & \text{if } t_{i,j}^{(w)} = 0, \\[6pt]
\begin{aligned}
&\bigl( 10 \cdot \mathds{1}[c_{i,j}^{(w)} = 1] 
+ 5 \cdot \mathds{1}[c_{i,j}^{(w)} = 2] \\
&\quad + \; \mathds{1}[c_{i,j}^{(w)} = 3] \bigr) 
\cdot (1 - u_{i,j})
\end{aligned}
& \text{if } t_{i,j}^{(w)} = 1.
\end{cases}
\]

This function specifies that the subregional outcome depends on whether a heat wave occurs and, if so, how vulnerable the subregion is based on the education level of parents and the vegetation coverage. If $t_{i,j}^{(w)}=0$, the effect is zero. If $t_{i,j}^{(w)}=1$, the impact is determined by the categorical education level with coefficients $(10,5,1)$ corresponding to low, medium, and high parental education, respectively, and is scaled by $(1 - u_{i,j})$, which reduces the effect in proportion to the amount of vegetation present. The indicator $\mathds{1}[\cdot]$ selects the appropriate coefficient for each subregion.

The regional outcome at month $w$ is obtained via mean aggregation. This setup induces a spatiotemporal causal problem with heterogeneous treatment effects driven by both observed and unobserved context, where the latter must be recovered from aggregated outcomes across multiple time points.

\paragraph{Training.}  
We train a model to jointly estimate:
\begin{enumerate}
    \item The mapping $f_{\boldsymbol{\theta}}(\cdot)$ from observed context, intervention, and vegetation index to subregional outcomes,
    \item The static vegetation index values ${U} \in [0,1]^{N \times M}$ for all subregions.
\end{enumerate}

The function $f_{\boldsymbol{\theta}}(\cdot)$ is implemented as a 4 layer multilayer perceptron (MLP) with hidden dimension $16$ and a dropout rate of $0.1$ after the second hidden layer. Its inputs for each subregion are:
\[
z_{i,j}^{(w)} = \big( \mathds{1}[c_{i,j}^{(w)} = 1], \ \mathds{1}[c_{i,j}^{(w)} = 2], \ \mathds{1}[c_{i,j}^{(w)} = 3], \ t_{i,j}^{(w)}, \ u_{i,j} \big),
\]
where $u_{i,j}$ is the vegetation index. The output is the predicted outcome $y_{i,j}^{(w)}$ for that subregion and month.

In the forward pass for a given month $w$:
\begin{enumerate}
    \item The model takes as input $T^{(w)}$, $C^{(w)}$, and ${U}$,
    \item The MLP evaluates $f_{\boldsymbol{\theta}}(\cdot)$ at each $(i,j)$ to produce predicted subregional outcomes $y_{i,j}^{(w)}$,
    \item These are aggregated to the regional level via mean aggregation.
    \item The primary loss is the MSE between predicted and observed regional outcomes.
\end{enumerate}

Training proceeds over all months in random order at each epoch. The total parameter set consists of the MLP weights $\boldsymbol{\theta}$ and the vegetation matrix ${U}$. We optimize using Adam with a learning rate of $0.001$ for $10{,}000$ epochs, with fixed random seeds for reproducibility. Loss curves for $\mathcal{L}_{\mathrm{region}}$ and the MSE of $\mathcal{L}_{\mathrm{veg}}$ are recorded throughout training.
\paragraph{Results.}  
\begin{enumerate}
    \item Figure~\ref{fig:results_exp3}a shows the loss curves and vegetation MSE under the linear scaling parameterization. Both training loss and MSE decrease rapidly, and the final error is very small, indicating near‐perfect recovery of the unobserved covariate.
    
    \item Figure~\ref{fig:results_exp3}b compares the ground‐truth vegetation field to the estimated vegetation under linear scaling. The two maps are almost indistinguishable, confirming that the restricted functional form matches the data-generating process and allows highly accurate recovery.
    
    \item Figures~\ref{fig:results_exp3}c and d present the same results for the non‐linear MLP parameterization. While the model still recovers meaningful structure in the vegetation, the MSE remains orders of magnitude higher than in the linear case, and deviations from the ground truth are clearly visible in the estimated field. This is notable since the true vegetation values are not necessarily identifiable; the MLP could instead store a transformed version and map them back during the forward process.
    
    \item Overall, these results demonstrate that the model can reconstruct regional outcomes while also uncovering hidden drivers of heterogeneity. The comparison between linear and non‐linear parameterizations highlights the trade‐off: structural restrictions yield more accurate recovery when they match the true mechanism, while flexible approximators such as MLPs still learn the vegetation but with reduced precision.
\end{enumerate}


\newpage

\section{Real-World Case Study Details: Effect of Heat on Gun Violence}
\label{app:realworld}

\begin{figure*}[t]
    \centering
    \includegraphics[width=0.8\linewidth]{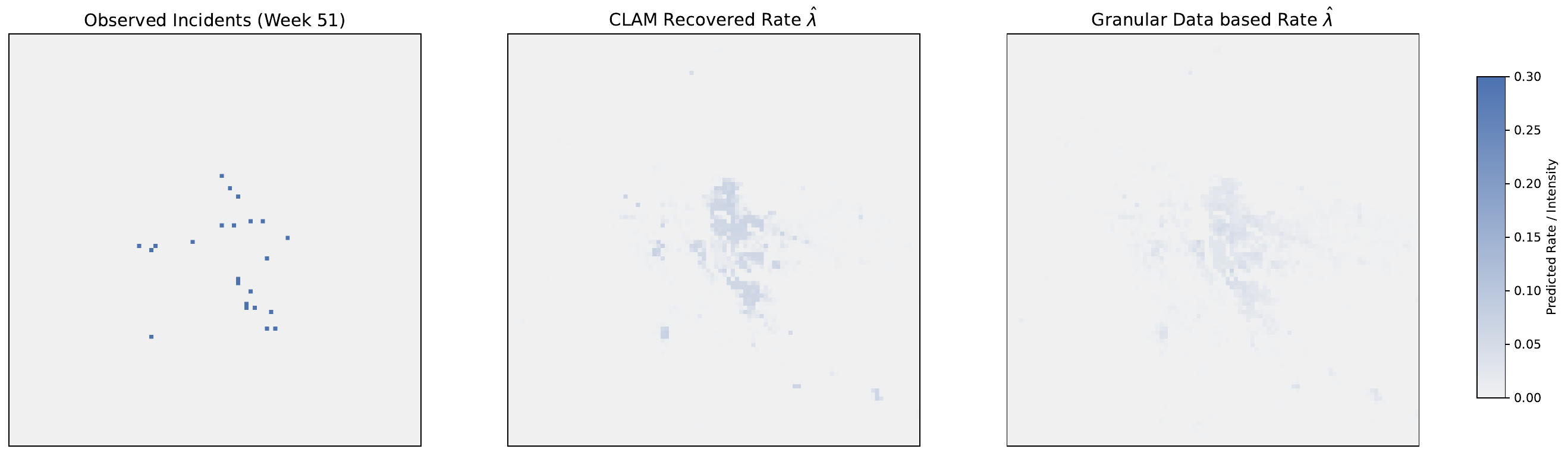}
    \caption{Comparison of learned spatial intensities for week~51. Both models are rendered on a shared color scale across the $\sqrt{N\cdot M}\times\sqrt{N\cdot M}$ grid, alongside the observed events for that week.}
    \label{fig:spatial_comparison_heatmap}
\end{figure*}

This section provides the full details of the real-world study summarized in Section~\ref{sec:real_world_experiment}. The notation follows Section~\ref{problemsetting}: regions are indexed by $i \in \{1,\dots,N\}$, subregions or cells by $j \in \{1,\dots,M\}$, and weeks by $w \in \{1,\dots,W\}$. Temperature is used as a continuous, time-varying treatment-like exposure; urbanization is the HR static contextual covariate; and gun-incident counts are the observed LR outcomes.

\paragraph{Problem Setting.}
The target variable is observed at a coarse spatial resolution, while covariates are available at a finer resolution. The study area is discretized into a fine grid of $\sqrt{N\cdot M} \times \sqrt{N\cdot M}$ cells, grouped into $\sqrt{N}\times\sqrt{N}$ coarse regions. With $N=100$ and $M=100$, this gives a $100\times 100$ cell grid grouped into a $10\times 10$ region grid. For each week $w$ and region $i$, we observe a nonnegative count outcome
\begin{equation}
Y^{\mathrm{LR}}_{i,w}\in\mathbb{Z}_{\ge 0},
\end{equation}
representing the number of gun-incident events occurring in region $i$ during week $w$.

\paragraph{Data Representation and Covariates.}
Let $c_{i,j,w}$ denote the observed urbanization measurement for cell $(i,j)$ and week $w$. Since urbanization is treated as time-invariant in this study, we use its temporal mean,
\begin{equation}
c_{i,j}
=
\frac{1}{W}\sum_{w=1}^{W} c_{i,j,w}.
\end{equation}
Temperature is modeled at the region level. Let $t^{\mathrm{HR}}_{i,j,w}$ denote the cell-level temperature for cell $(i,j)$ and week $w$. We compute the region-average temperature exposure as
\begin{equation}
t^{\mathrm{LR}}_{i,w}
=
\frac{1}{M}\sum_{j=1}^{M} t^{\mathrm{HR}}_{i,j,w},
\end{equation}
and standardize it using the global mean and standard deviation across all region--week pairs:
\begin{equation}
\widetilde{t}_{i,w}
=
\frac{t^{\mathrm{LR}}_{i,w}-\mu_t}{\sigma_t+\varepsilon},
\end{equation}
where $\varepsilon>0$ is a small constant for numerical stability. Missing temperature values are imputed using the global median temperature, and missing outcome counts are set to zero.

\paragraph{Disaggregation Model.}
We model the latent HR intensity, or rate, $\lambda_{i,j,w}>0$ as a function of the standardized regional temperature exposure and the HR urbanization covariate:
\begin{equation}
\lambda_{i,j,w}
=
f_{\boldsymbol{\theta}}\!\left(\widetilde{t}_{i,w},\, c_{i,j}\right),
\label{eq:cell_rate}
\end{equation}
where $f_{\boldsymbol{\theta}}(\cdot)$ is a feed-forward neural network with parameters $\boldsymbol{\theta}$. To enforce nonnegativity, the network output is passed through a softplus transformation.

The model-implied LR expected count is obtained by summing the HR rates over all cells in region $i$:
\begin{equation}
\mu_{i,w}
=
\sum_{j=1}^{M}\lambda_{i,j,w}.
\label{eq:region_sum}
\end{equation}
Thus, $\mu_{i,w}$ denotes the predicted regional count intensity, while $Y^{\mathrm{LR}}_{i,w}$ denotes the observed regional count. This additive aggregation is consistent with the SCM in Section~\ref{sec:scmmain}.

\paragraph{Training Objective.}
We assume a Poisson observation model at the region--week level,
\begin{equation}
Y^{\mathrm{LR}}_{i,w}
\sim
\mathrm{Poisson}\!\left(\mu_{i,w}\right),
\end{equation}
and train $f_{\boldsymbol{\theta}}$ by minimizing the negative Poisson log-likelihood, up to additive constants:
\begin{equation}
\mathcal{L}(\boldsymbol{\theta})
=
\frac{1}{|\mathcal{W}_\mathrm{train}|\cdot N}
\sum_{w\in\mathcal{W}_\mathrm{train}}
\sum_{i=1}^{N}
\left(
\mu_{i,w}
-
Y^{\mathrm{LR}}_{i,w}\log(\mu_{i,w}+\varepsilon)
\right),
\label{eq:poisson_nll}
\end{equation}
where $\mathcal{W}_\mathrm{train}$ is the set of training weeks and $\varepsilon>0$ is a small constant for numerical stability.

\paragraph{Spatial Intensity Comparison.}
Figure~\ref{fig:spatial_comparison_heatmap} visualizes observed incidents for a representative week, week~51, alongside spatial intensity estimates learned under different supervision regimes. When trained only on region-level counts, \textsc{Clam} produces spatially structured intensity maps that concentrate mass around incident clusters but exhibit noticeably higher peak intensities and broader spatial support compared to the cell-supervised estimate. This overestimation reflects a fundamental ambiguity induced by aggregation: without access to fine-grained outcomes, the model must allocate regional mass across plausible subregional locations, leading to inflated local rates in areas consistent with the observed context and covariates. The response-surface comparison in the main paper, Figure~\ref{fig:comparison_heatmap}, shows the same pattern in covariate space rather than physical space.

\paragraph{Learning.}
The training procedure optimizes the function parameters \(\boldsymbol{\theta}\) (or any other missing components of the SCM).  
The structural function \(f_{\boldsymbol{\theta}}(\cdot)\) can be represented by a neural network or any other parameterized function.  
Since the fine-grained outcomes \(y_{i,j}\) are unobserved, training relies solely on their aggregated counterparts.

We optimize with the goal that the inferred aggregated predictions ($\mu_i$) match the observed aggregate:
$
\mathcal{L}_{\boldsymbol{\theta},\boldsymbol{\phi}} =
\sum_{i=1}^{N}
\bigl(
\widehat{Y}_i - \mu_i
\bigr)^{2},
$
where $\widehat{Y}_i$ denotes the observed aggregate of region $i$.
If the noise terms are not i.i.d. and normally distributed, they can also be explicitly learned within the optimization loop.

\newpage
\section{Additional Experiments}
\label{sec:addexp}
Exp.~4 considers driving bans with an unknown aggregation function between subregional and regional outcomes. 
Exp.~5 introduces treatment–context confounding, where the probability of subregional treatment depends on contextual variables, affecting both allocation and outcomes

\subsection{Exp.~4: Unknown Aggregation Functions}
\begin{figure}
    \centering
    \includegraphics[width=0.97\linewidth]{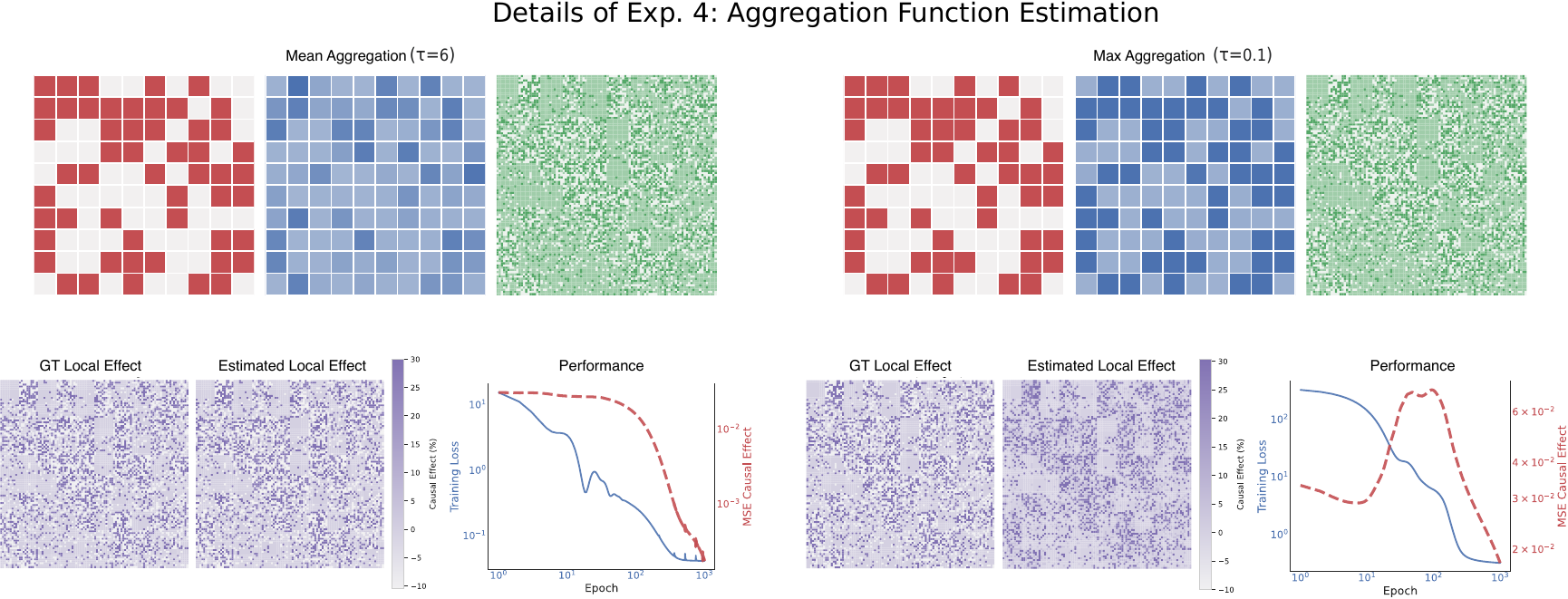}
    \caption{\textbf{Results Overview.} 
    \textbf{Top:} Input of LR intervention locations (\textcolor[rgb]{0.768,0.305,0.321}{red}), 
    LR outcomes (\textcolor[rgb]{0.298,0.447,0.690}{blue}), 
    and HR context (\textcolor[rgb]{0.333,0.658,0.407}{green}) for mean aggregation and max aggregation cases. 
    \textbf{Bottom:} We observe the estimated local effects and the performance of our model over multiple epochs for both cases.}
    \label{fig:exp4}
\end{figure}

\paragraph{Context.}
We study the causal effect of driving bans on air quality under the assumption that the reported region-level air quality values are an unknown aggregation of finer-scale (subregional) measurements. Specifically, we model the aggregation as a learned combination of the subregional mean and maximum values, with the combination weights parameterized via a logit transform.

A realistic example is a city-wide driving ban where air quality is officially reported as a single regional index (e.g., PM$_{2.5}$ concentration), while the underlying measurement network collects data from multiple monitoring stations. Depending on reporting policies, the published index might resemble an average across stations, a worst-case station reading, or something in between.

The high-resolution context for each subregion is a single continuous variable: a vegetation index, capturing the amount of green coverage in the area. Vegetation can mitigate pollution and thus may modulate the effect of a driving ban. We hypothesize that the causal effect of driving bans varies with vegetation and that the unknown aggregation mechanism must be learned alongside the effect parameters to estimate subregional impacts correctly.

\paragraph{Setup.}  
We assume $N = 100$ regions, each with $M = 100$ subregions.  

We are given:
\begin{itemize}
    \item A binary treatment vector $T \in \{0,1\}^N$, where $T_i = 1$ indicates that a driving ban was implemented in region $i$ (and thus all its subregions).
    \item A high-resolution context matrix $C \in [0,1]^{N \times M}$ encoding the vegetation index of each subregion, where $c_{i,j} = 0$ represents minimal vegetation cover and $c_{i,j} = 1$ represents dense vegetation.
    \item A high-resolution noise matrix $\epsilon \in \mathbb{R}^{N \times M}$ with i.i.d.\ entries $\epsilon_{i,j} \sim \mathcal{N}(0, \sigma^2)$, where $\sigma^2 = 0.02$.
\end{itemize}

The treatment vector $T$ is generated using a Bernoulli distribution with probability $0.5$, assigning each region to either the intervention or control group at random. The vegetation index values in $C$ are drawn i.i.d.\ from $\mathrm{Uniform}(0,1)$.

The subregional outcome variable is given by the ground-truth functional relationship:
\[
y_{i,j} = f_{\boldsymbol{\theta}}\!\left(t_{i,j}, c_{i,j}\right) + \epsilon_{i,j},
\]
where:
\begin{itemize}
    \item $t_{i,j} = T_i$ is the treatment status of subregion $j$ in region $i$ (inherited from the region),
    \item $c_{i,j}$ is the vegetation index,
    \item $\epsilon_{i,j}$ is Gaussian noise.
\end{itemize}

The causal effect function is parameterized as:
\[
f_{\boldsymbol{\theta}}(t_{i,j}, c_{i,j}) =
\begin{cases}
0.0 & \text{if } t_{i,j} = 0, \\
\theta_{\mathrm{base}} + \theta_{\mathrm{veg}} \cdot c_{i,j} & \text{if } t_{i,j} = 1,
\end{cases}
\]
where $\theta_{\mathrm{base}}$ captures the baseline effect of a driving ban and $\theta_{\mathrm{veg}}$ captures how this effect changes with vegetation cover.

Unlike Exp.~1, the regional outcome $x_i$ is not a simple mean over subregions.  
Instead, we model the aggregation function as:
\[
y_i = \sum_{j=1}^M p_{i,j}(\tau) \cdot y_{i,j},
\]
where:
\[
p_{i,j}(\tau) = \frac{\exp\!\left( y_{i,j} / \tau \right)}{\sum_{k=1}^M \exp\!\left( y_{i,k} / \tau \right)}
\]
is the softmax weight assigned to subregion $j$ with temperature parameter $\tau > 0$.  
When $\tau \to \infty$, $p_{i,j}$ approaches a uniform distribution (mean aggregation); when $\tau \to 0$, $p_{i,j}$ concentrates on the subregion with the highest outcome (max aggregation).

\paragraph{Training.}  
We train a model to estimate:
\begin{enumerate}
    \item The parameters $\boldsymbol{\theta} = \{\theta_{\mathrm{base}}, \theta_{\mathrm{veg}}\}$ of the causal effect function $f_{\boldsymbol{\theta}}(\cdot)$,
    \item The aggregation temperature parameter $\tau > 0$ governing the softmax-based aggregation from subregional to regional outcomes.
\end{enumerate}

The learned function takes the form:
\[
f_{\boldsymbol{\theta}}(t_{i,j}, c_{i,j}) =
\begin{cases}
0.0 & \text{if } t_{i,j} = 0, \\
\theta_{\mathrm{base}} + \theta_{\mathrm{veg}} \cdot c_{i,j} & \text{if } t_{i,j} = 1.
\end{cases}
\]

In the forward pass:
\begin{enumerate}
    \item Compute predicted subregional outcomes:
    \(
    y_{i,j} = f_{\boldsymbol{\theta}}\!\left(t_{i,j}, c_{i,j}\right),
    \)
    \item Compute subregion-level aggregation weights via the softmax:
    \(
    p_{i,j}(\tau) = \frac{\exp\!\left( y_{i,j} / \tau \right)}{\sum_{k=1}^M \exp\!\left( y_{i,k} / \tau \right)},
    \)
    \item Compute predicted regional outcomes as the expectation under $p_{i,j}(\tau)$:
    \(
    y_i = \sum_{j=1}^M p_{i,j}(\tau) \cdot y_{i,j},
    \)
    \item Compute the loss as the mean squared error with the observed $Y_i$.
\end{enumerate}

All parameters $\{\theta_{\mathrm{base}}, \theta_{\mathrm{veg}}, \tau\}$ are optimized jointly using Adam with a learning rate of $0.001$ for $1000$ epochs. We initialize $\tau$ to a moderate value (e.g., $\tau=1.0$) and enforce $\tau > 0$ during training by optimizing its logarithm.

\paragraph{Results.}  
\begin{enumerate}
    \item Visualizations of the aggregated causal effects show that mean aggregation preserves more variation across regions, whereas max aggregation suppresses smaller effects and produces sharper contrasts.
    \item Although the subregional causal effect loss is never used directly for training, it decreases steadily over epochs, indicating that the model recovers fine scale effects from only regional supervision. Training loss is substantially higher under max aggregation, as weaker subregional signals are obscured.
    \item Comparing estimated and ground truth local effects confirms this bias: areas with small true effects tend to be overestimated when max aggregation dominates.
    \item The aggregation function, parameterized by the temperature $\tau$, is recovered with high accuracy. In the mean aggregation case, $\tau$ is estimated within $0.01$ of the ground truth, while in the max aggregation case, the estimate is slightly underestimated ($5.6$), consistent with the plateauing behavior of the softmax function at low temperatures.
    \item Overall, the experiment demonstrates that our approach can successfully recover both the aggregation mechanism (within its parametric form) and the underlying causal effects, despite only observing region-level outcomes.
\end{enumerate}

\subsection{Exp.~5: Covariate-Based Confounding of Treatment Subregion}
\begin{figure}
    \centering
    \includegraphics[width=0.9\linewidth]{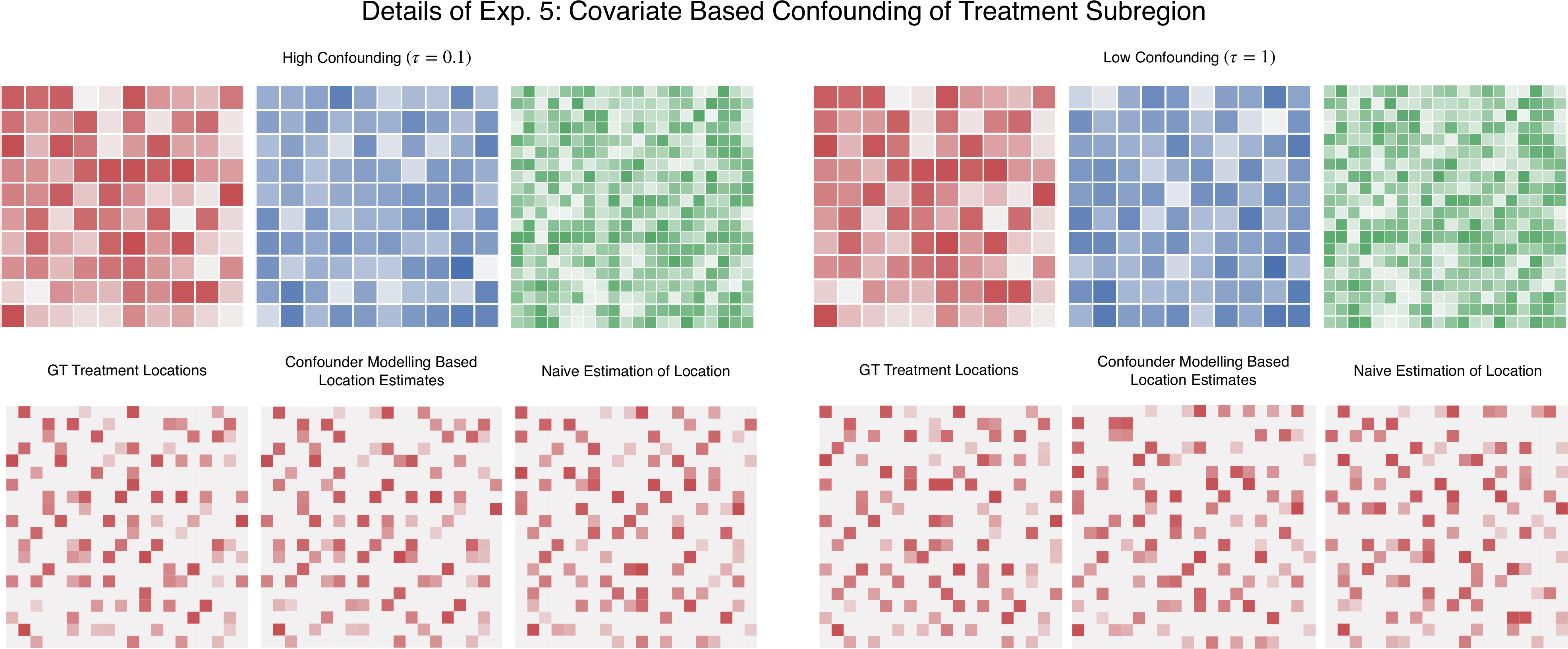}
    \caption{\textbf{Results Overview.} 
    \textbf{Top:} Input of LR intervention locations (\textcolor[rgb]{0.768,0.305,0.321}{red}), 
    LR outcomes (\textcolor[rgb]{0.298,0.447,0.690}{blue}), 
    and HR context (\textcolor[rgb]{0.333,0.658,0.407}{green}). 
    \textbf{Bottom:} Ground truth treatment locations, estimation of treatment locations based on modeling of location covariate-based confounding, naïve estimation of treatment locations with all free parameters.}
    \label{fig:results_exp5}
\end{figure}
\paragraph{Context.}  
In this experiment, we study how confounding between treatment allocation and contextual variables affects causal effect estimation. We take the example of school districts receiving additional public funding and the corresponding changes in educational outcomes.  

In reality, funding allocation is rarely random: wealthier or poorer districts may have systematically different chances of receiving funding due to political priorities, lobbying power, or policy constraints. This creates a dependency between the probability of treatment and district level context variables, violating the standard assumption of independent treatment assignment.

We model this confounding explicitly by making the probability of treatment allocation a deterministic function of the contextual variable, passed through a logistic transform to obtain valid probabilities for each region. The treatment effect itself is also a function of the same contextual variable, representing heterogeneous impacts across different socioeconomic conditions.

\paragraph{Setup.}
We assume $N = 100$ regions, each with $M = 100$ subregions (school districts).

We are given:
\begin{itemize}
    \item A high-resolution treatment matrix $T \in \{0,1\}^{N \times M}$ with \emph{exactly one} treated subregion per region, i.e., $\sum_{j=1}^{M} t_{i,j} = 1$ for all $i$.
    \item A context matrix $C \in [0,1]^{N \times M}$ encoding a socioeconomic/need index for each subregion, where a higher $c_{i,j}$ indicates greater need.
    \item A noise matrix $\epsilon \in \mathbb{R}^{N \times M}$ with i.i.d.\ entries $\epsilon_{i,j} \sim \mathcal{N}(0,\sigma^2)$ (default $\sigma^2 = 0.02$).
\end{itemize}

\paragraph{Confounded Treatment Allocation.}
For each region $i$, we form logits from the subregional context,
\[
\ell_{i,j} = \alpha_0 + \alpha_1\, c_{i,j},
\]
and convert them to a categorical distribution over subregions via a row-wise softmax,
\[
p_{i,j} \;=\; \frac{\exp(\ell_{i,j})}{\sum_{k=1}^{M} \exp(\ell_{i,k})}.
\]
We then draw exactly one treated subregion $j^\star \sim \mathrm{Categorical}(p_{i,1:M})$ and set
\[
t_{i,j} = \mathds{1}\{j=j^\star\}, \qquad \sum_{j=1}^{M} t_{i,j} = 1.
\]
The parameters $(\alpha_0,\alpha_1)$ control the strength and direction of confounding between context and treatment assignment (default $\alpha_0=0$, $\alpha_1>0$ so higher-need subregions are more likely to be treated).

\paragraph{Outcome Model With Context-Dependent Effects.}
Subregional outcomes follow the ground-truth functional relationship
\[
y_{i,j} \;=\; f_{\boldsymbol{\theta}}\!\left(t_{i,j}, c_{i,j}\right) + \epsilon_{i,j},
\]
with the causal effect function
\[
f_{\boldsymbol{\theta}}(t_{i,j}, c_{i,j}) \;=\;
\begin{cases}
0 & \text{if } t_{i,j}=0,\\[2pt]
\theta_{\mathrm{base}} + \theta_{\mathrm{soc}}\, c_{i,j} & \text{if } t_{i,j}=1,
\end{cases}
\]
where $\boldsymbol{\theta}=\{\theta_{\mathrm{base}},\,\theta_{\mathrm{soc}}\}$ captures the baseline funding effect and its modulation by socioeconomic need.

\paragraph{Regional Aggregation.}
Region-level outcomes are the sum over subregions (as in Exp.~1).

This construction induces confounding because (i) treatment allocation depends on $c_{i,j}$ through the softmax logits, and (ii) treatment effects also vary with $c_{i,j}$ via $f_{\boldsymbol{\theta}}(\cdot)$.

\paragraph{Training.}  
We compare two training regimes for estimating the causal effect parameters $\boldsymbol{\theta} = \{\theta_{\mathrm{base}}, \theta_{\mathrm{soc}}\}$ and the high resolution treatment assignments ${T} \in \mathbb{R}^{N \times M}$ in the presence of treatment context confounding.

\paragraph{Regime~1: Naïve Training Under Independence Assumption.}  
This setup directly applies the training procedure of Exp.~2 (Section~\ref{app:expdetails}) to the confounded data, it:
\begin{enumerate}
    \item Treats the treatment allocation as if independent of the context,
    \item Learns ${T}$ as free parameters, constrained to be one hot per row via a differentiable argmax with temperature, 
    \item Learns $\boldsymbol{\theta}$ by minimizing the MSE between observed and predicted regional outcomes.
\end{enumerate}
This ignores the fact that the treatment assignment mechanism is context dependent.

\paragraph{Regime~2: Confounding Aware Training.}  
Here, we explicitly model the context-dependent treatment allocation mechanism.  
For each region $i$, we predict logits 
\[
\ell_{i,j} = \beta_0 + \beta_1 \, c_{i,j},
\]
where $(\beta_0, \beta_1)$ are learned parameters, and convert them to a categorical distribution via a row-wise softmax with a learnable temperature $\tau_{\mathrm{conf}} > 0$:
\[
p_{i,j}(\tau_{\mathrm{conf}}) = \frac{\exp\!\left( \ell_{i,j} / \tau_{\mathrm{conf}} \right)}{\sum_{k=1}^M \exp\!\left( \ell_{i,k} / \tau_{\mathrm{conf}} \right)}.
\]
We then use $p_{i,j}(\tau_{\mathrm{conf}})$ as the (differentiable) treatment assignment in the forward pass.

In both regimes, predicted subregional outcomes are given by:
\[
y_{i,j} = f_{\boldsymbol{\theta}}\!\left(t_{i,j}, c_{i,j}\right)
= \begin{cases}
0 & \text{if } t_{i,j} = 0, \\[2pt]
\theta_{\mathrm{base}} + \theta_{\mathrm{soc}} \cdot c_{i,j} & \text{if } t_{i,j} = 1,
\end{cases}
\]
where in Regime~1, $t_{i,j}$ comes from the naïvely learned $\widehat{T}$, and in Regime~2, $t_{i,j} = p_{i,j}(\tau_{\mathrm{conf}})$ from the confounding model.

Regional outcomes are summed and aggregated.
and the training loss is the mean squared error,
All parameters are optimized jointly using Adam with a learning rate of $0.001$ for $1000$ epochs.  
Regime~1 learns $\{\widehat{T}, \theta_{\mathrm{base}}, \theta_{\mathrm{soc}}\}$,  
while Regime~2 learns $\{\beta_0, \beta_1, \tau_{\mathrm{conf}}, \theta_{\mathrm{base}}, \theta_{\mathrm{soc}}\}$.

\paragraph{Results.}  
\begin{enumerate}
    \item We compare two cases of confounding in subregional treatment allocation, holding region-level treatments and contextual factors fixed. The aggregated causal effects differ markedly depending on the level of confounding, showing that the treatment context dependency directly shapes the observed regional outcomes.
    
    \item We visualize the ground truth intervention locations alongside the estimates obtained with and without explicitly modeling confounding. Under strong confounding, the confounding-aware approach recovers the true intervention locations with high fidelity, while the naïve method from Exp.~2 retrieves only a small fraction correctly.
    
    \item In the low confounding setting, both methods perform similarly, and location probabilities are estimated with comparable accuracy. This confirms that adjusting for confounding does not reduce performance when confounding is weak.
    
    \item For causal effect estimation, the final MSE of subregional effects under low confounding was $2.04$ (naïve) versus $1.60$ (confounding aware). Under high confounding, the respective errors were $0.74$ versus $0.32$. These results show that accounting for confounding improves both intervention location recovery and the accuracy of estimated causal effects by conditioning on the correct contextual information.
\end{enumerate}

Overall, this experiment demonstrates that explicitly modeling treatment context confounding is essential as it enables accurate recovery of intervention locations and yields more reliable causal effect estimates, particularly when treatment assignment is strongly biased by contextual factors.

\end{document}